\documentclass{article}
\usepackage{graphicx}
\usepackage{subcaption}
\usepackage{xurl}
\usepackage{tikz}
\usetikzlibrary{calc}
\usetikzlibrary{fit}
\usetikzlibrary{patterns}
\usetikzlibrary{decorations.pathreplacing}
\usetikzlibrary{arrows,calc,arrows.meta}
\usetikzlibrary{matrix,positioning}
\usetikzlibrary{shapes.geometric}
\PassOptionsToPackage{numbers, compress}{natbib}
\usepackage[preprint]{neurips_2026}

\usepackage[utf8]{inputenc} 
\usepackage[T1]{fontenc}    
\usepackage{amsmath}
\usepackage{hyperref}
\usepackage[capitalise,nameinlink]{cleveref}
\usepackage{url}            
\usepackage{booktabs}       
\usepackage{amsfonts}       
\usepackage{nicefrac}       
\usepackage{microtype}      
\usepackage{wrapfig}
\usepackage{comment}

\usepackage{amssymb}
\usepackage{mathtools}
\usepackage{amsthm}
\usepackage{pgfplots}
\usepackage{pgfplotstable}
\usepackage{float}
\usepackage{enumitem}
\usepackage{fontawesome5}

\usepackage{xcolor}
\usepackage[most]{tcolorbox}
\definecolor{googleblue}   {HTML}{4285F4}
\definecolor{googlered}    {HTML}{EA4335}
\definecolor{googleyellow} {HTML}{FBBC05}
\definecolor{googlegreen}  {HTML}{34A853}
\definecolor{googlegray}      {HTML}{5F6368}

\usepackage{colortbl}
\definecolor{LightGray}{gray}{0.9}

\newcommand\blfootnote[1]{%
  \begingroup
  \renewcommand\thefootnote{}%
  \footnotetext{#1}%
  \addtocounter{footnote}{-1}%
  \endgroup
}
\title{Spreadsheet-RL: Advancing Large Language Model Agents on Realistic Spreadsheet Tasks via Reinforcement Learning}

\author{%
Banghao Chi$^{1*}$, Yining Xie$^{1*}$,  Mingyuan Wu$^{1*\dagger}$,
\bf Jingcheng Yang$^{1}$, Jize Jiang$^{1}$, Zhaoheng Li$^{1}$, \\
 \bf Shengyi Qian$^{2}$, Minjia Zhang$^{1}$, Klara Nahrstedt$^{1}$, Rui Hou$^{2}$, Xiangjun Fan$^{2}$, Hanchao Yu$^{2}$ \\
$^{1}$University of Illinois Urbana-Champaign,
$^{2}$Meta \\
\texttt{\{banghao2, yining19, mw34\}@illinois.edu,} \\
}

\begin{document}
  \noindent\begin{tabular*}{\textwidth}{@{}l@{\extracolsep{\fill}}r@{}}
    \raisebox{-0.15\height}{\includegraphics[height=0.8cm]{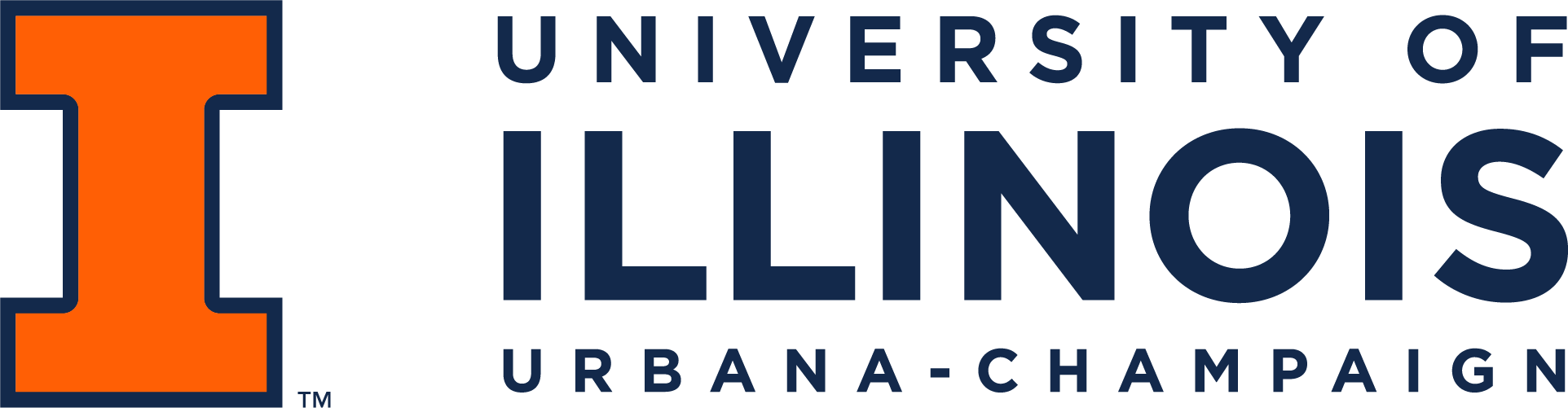}}\hspace{0.6em}%
    \raisebox{-0.15\height}{\includegraphics[height=0.7cm]{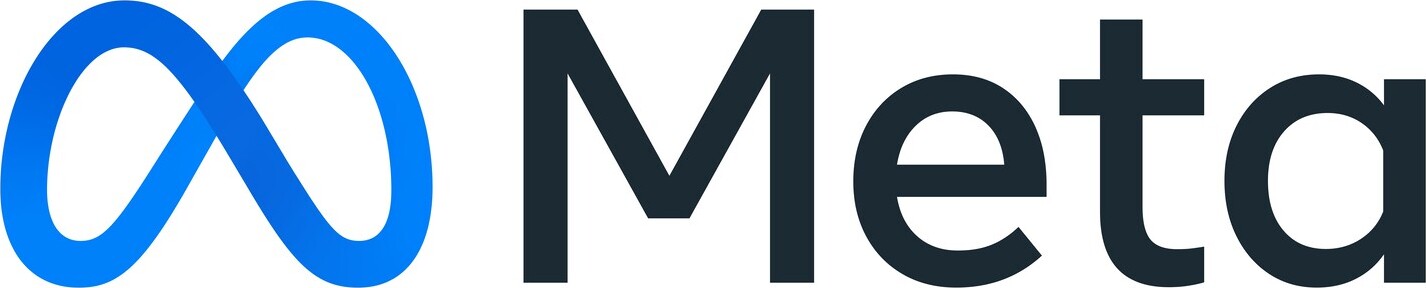}} &
    \raisebox{-0.15\height}{\includegraphics[height=1cm]{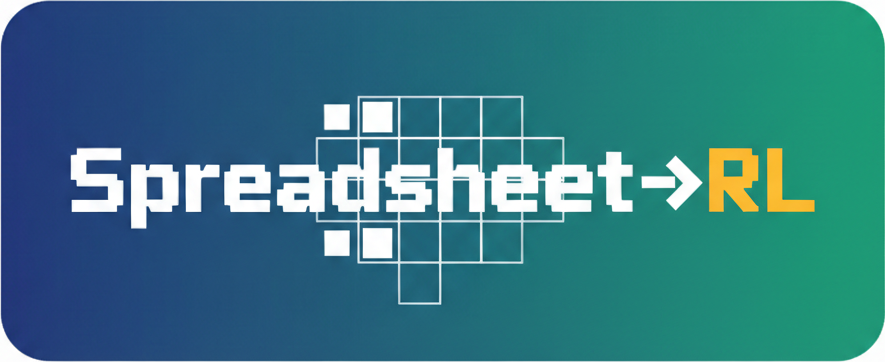}}
  \end{tabular*}

  \vspace{0.2cm}

\maketitle
\blfootnote{All data and code releases are maintained by the corresponding authors at UIUC and are not affiliated with Meta.}
\blfootnote{* Contributed equally to this work, where more junior authors are listed ahead of senior.}
\blfootnote{$^\dagger$ Project Lead.}
\vspace{-2.5em}
\begin{center}
\setlength{\tabcolsep}{5pt}
\renewcommand{\arraystretch}{1.05}
\begin{tabular}{@{}c l@{}}
\raisebox{-0.35\height}{\includegraphics[height=1.7em]{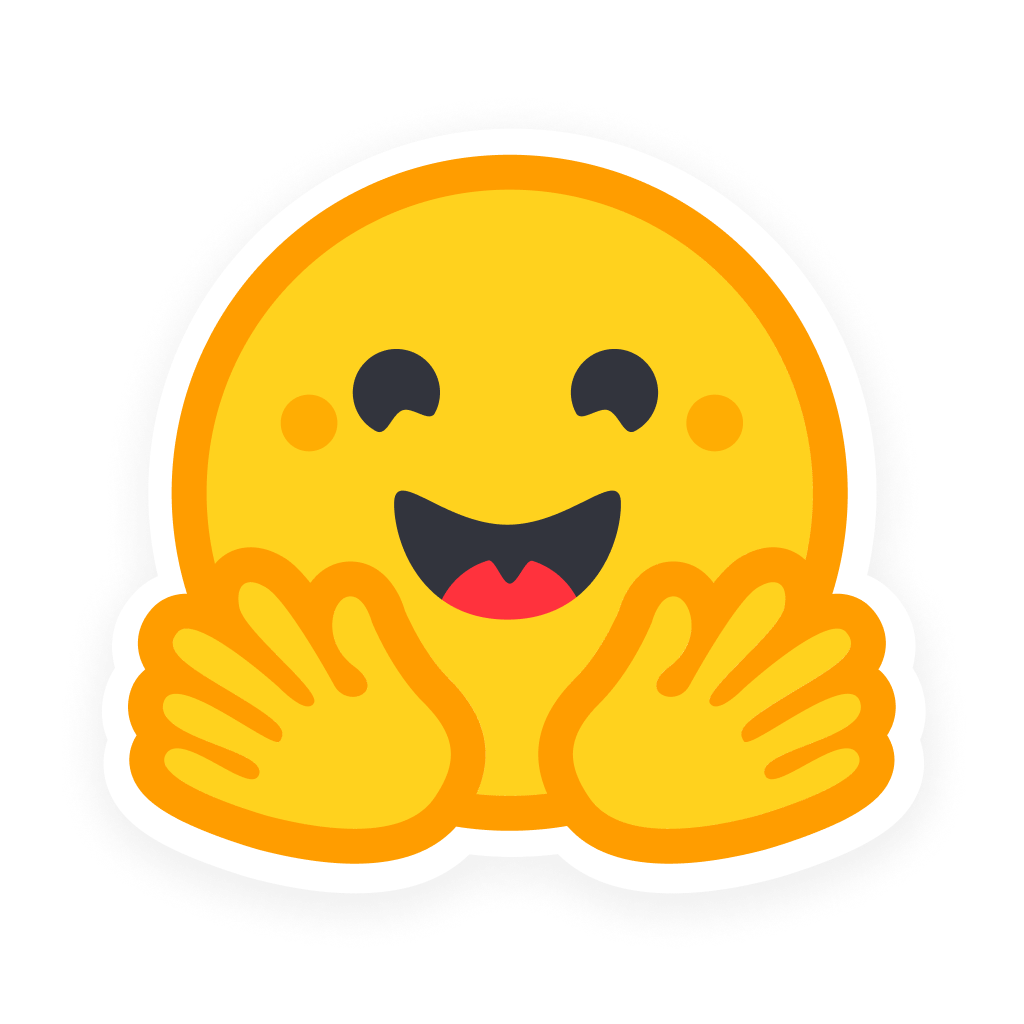}} &
\href{https://huggingface.co/datasets/Spreadsheet-RL/Spreadsheet-RL}
{\nolinkurl{https://huggingface.co/datasets/Spreadsheet-RL/Spreadsheet-RL}} \\

\raisebox{-0.2\height}{\scalebox{1.3}{\faGithub}} &
\href{https://github.com/Spreadsheet-RL/Spreadsheet-RL}
{\nolinkurl{https://github.com/Spreadsheet-RL/Spreadsheet-RL}} \\
\end{tabular}
\end{center}
\vspace{-0.5em}

\begin{abstract}

Spreadsheet systems (e.g., Microsoft Excel, Google Sheets) play a central role in modern data-centric workflows. As AI agents grow increasingly capable of automating complex tasks, such as controlling computers and generating presentations, building an AI-driven spreadsheet agent has emerged as a promising research direction. Most existing spreadsheet agents rely on specialized prompting over general-purpose LLMs; while this design has potentials on simple spreadsheet operations, it struggles to manage the complex, multi-step workflows typical of real-world applications.

In this paper, we introduce \textbf{Spreadsheet-RL}, a reinforcement learning (RL) fine-tuning framework designed to train specialized spreadsheet agents within a realistic Microsoft Excel environment. \textbf{Spreadsheet-RL} features an automated pipeline for scalable collection of paired start-goal spreadsheets from online forums, as well as domain-specific evaluation tasks in areas such as finance and supply chain management, which we compile into the new \textbf{Domain-Spreadsheet} benchmark dataset. It also includes a \textbf{Spreadsheet Gym} environment designed for multi-turn RL: \textbf{Spreadsheet Gym} exposes extensive Excel functionality through a Python sandbox, along with a refined harness that incorporates a comprehensive tool set and carefully designed tool-routing rules for spreadsheet tasks. Through comprehensive experiments, we show that \textbf{Spreadsheet-RL} substantially enhances AI agent's performance on both general and domain-specific spreadsheet tasks: it improves Qwen3-4B-Thinking-2507's Pass@1 on SpreadsheetBench from \textbf{12.0\%} to \textbf{23.4\%}, and raises Pass@1 from \textbf{8.4\%} to \textbf{17.2\%} on our curated Domain-Spreadsheet dataset. These results highlight \textbf{Spreadsheet-RL}'s strong potential for generalization and real-world adoption in spreadsheet automation, and broadly, its promise for advancing LLM-based interactions with data interfaces in everyday work. We will release the training data, environment, and training pipeline to facilitate future research on spreadsheet agents.
\end{abstract}

\section{Introduction}
Spreadsheet systems, such as Microsoft Excel, Google Sheets, WPS Sheets, and LibreOffice, are widely adopted in data-centric workflows~\cite{de2022toward, benchmark_spreadsheet}. They support tasks from \textit{personal} activities such as travel planning and household budgeting, to \textit{professional} duties such as financial modeling and data presentation~\citep{benchmark_spreadsheet, spreadsheet_mot,visual}. As AI agents grow in both popularity and capability for automating tasks traditionally performed by humans, such as computer use and slide deck design~\citep{wang2025aiagentshumanwork, koh-etal-2024-visualwebarena,ge2025autopresentdesigningstructuredvisuals,osworld}, the development of an AI agent for spreadsheets (a \textit{spreadsheet agent}) to automate the human-operated spreadsheet-centered workflows will have the potential to fundamentally reshape how data science is performed at scale.

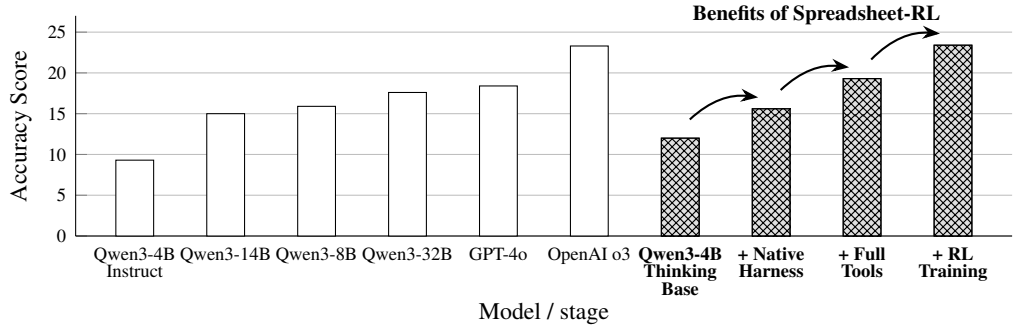
\begin{figure}[t]
\centering
\usetikzlibrary{patterns}

\begin{tikzpicture}
\begin{axis}[
    ybar,
    clip=false,
    width=\linewidth,
    height=45mm,
    bar width=5mm,
    ymin=0,
    ymax=27,
    ylabel style={yshift=-3.2ex, align=center},
    xlabel style={yshift=-1.5ex},
    axis y line*=none,
    axis x line*=none,
    ytick={0, 5, 10, 15, 20, 25},
    yticklabels={0, 5, 10, 15, 20, 25},
    xtick={1, 2, 3, 4, 5, 6, 7, 8, 9, 10},
    xtick style={draw=none},
    xticklabels={
        Qwen3-4B\\[-0.2em]Instruct,
        Qwen3-14B,
        Qwen3-8B,
        Qwen3-32B,
        GPT-4o,
        OpenAI o3,
        \textbf{Qwen3-4B}\\[-0.2em]\textbf{Thinking}\\[-0.2em]\textbf{Base},
        \textbf{+ Native}\\[-0.2em]\textbf{Harness},
        \textbf{+ Full}\\[-0.2em]\textbf{Tools},
        \textbf{+ RL}\\[-0.2em]\textbf{Training}
    },
    x tick label style={yshift=0.8ex},
    xmin=0.35,
    xmax=10.65,
    ymajorgrids,
    tick label style={font=\scriptsize, align=center},
    label style={font=\small},
    ylabel={Accuracy Score},
    xlabel={Model / stage},
    area legend,
    bar shift=0pt,
]

\addplot[black, fill=white] coordinates {(1,9.3)};
\addplot[black, fill=white] coordinates {(2,15.0)};
\addplot[black, fill=white] coordinates {(3,15.9)};
\addplot[black, fill=white] coordinates {(4,17.6)};
\addplot[black, fill=white] coordinates {(5,18.4)};
\addplot[black, fill=white] coordinates {(6,23.3)};
\addplot[black, fill=gray!25, postaction={pattern=crosshatch}] coordinates {(7,12.0)};
\addplot[black, fill=gray!25, postaction={pattern=crosshatch}] coordinates {(8,15.6)};
\addplot[black, fill=gray!25, postaction={pattern=crosshatch}] coordinates {(9,19.3)};
\addplot[black, fill=gray!25, postaction={pattern=crosshatch}] coordinates {(10,23.4)};

\begin{scope}[
    >={Stealth[black]},
    every node/.style={fill=none,circle},
    every edge/.style={draw=black, thick}
]
\node[] (a) at (axis cs:7,13.2) {};
\node[] (b) at (axis cs:8,16.8) {};
\node[] (c) at (axis cs:9,20.5) {};
\node[] (d) at (axis cs:10,24.7) {};
\draw[->, thick] (a) to[bend left=25] (b);
\draw[->, thick] (b) to[bend left=25] (c);
\draw[->, thick] (c) to[bend left=25] (d);
\end{scope}

\node[
    anchor=south,
    black,
    align=center,
    font=\footnotesize\bfseries
] at (axis cs:8.5,24.8)
{Benefits of Spreadsheet-RL};

\end{axis}
\end{tikzpicture}

\caption{
The rightmost four highlighted bars trace the main Qwen3-4B-Thinking-2507 result sequence: raw base model, spreadsheet-native interaction harness, comprehensive spreadsheet-tool access, and Spreadsheet-RL post-training. Representative open-source and closed-sourced  baselines are provided for scale reference.
}
\label{fig:intro}
\end{figure}

While there exists recent research such as SheetCopilot~\citep{sheetcopilot}, SheetAgent~\citep{sheetagent}, and ChatGPT Agent~\citep{openaiagent} studying spreadsheet agents, their approaches notably rely on proprietary (and powerful) Large Language Models (LLMs) with reasoning such as GPT-4o~\citep{openai2024gpt4ocard}, which have sufficient general capabilities to perform simple spreadsheet operations via natural language instructions. That is, these works are limited in that they rely on advancements in general LLMs, and their prompting strategies, rather than specific improvements in how the LLM agents utilize spreadsheets.
This limitation leads to existing spreadsheet agents struggling to reliably execute more complex, multi-step workflows that dominate real-world spreadsheet use; for example, the ChatGPT Agent and Copilot (both with excel access)~\citep{openaiagent}, reach 45.5\% and 20.0\%, respectively, on SpreadsheetBench~\citep{spreadsheetbench}.
On the other hand, frontier industry labs have recently begun to develop specialized spreadsheet agents, yet adopt undisclosed approaches that rely on internal benchmarks and closed training pipelines~\citep{openai2025introducinggpt52, microsoft, google}.

One promising approach a powerful, specialized, and open-source spreadsheet agent can be built is reinforcement learning (RL) fine-tuning: following DeepSeek-R1~\citep{Guo_2025}, on-policy RL combined with rule-based, verifiable outcome rewards has improved mathematical reasoning~\citep{shao2024deepseekmathpushinglimitsmathematical,wang2025reinforcementlearningreasoninglarge}, visual reasoning~\citep{wu2026vtoolr1vlmslearnthink}, and enabled scalable post-training for agentic domains such as software engineering~\citep{wei2025swerladvancingllmreasoning,wei2025trainingsuperintelligentsoftwareagents}, web interaction~\citep{bai2026webgymscalingtrainingenvironments}, data \citep{song2026agentdataprotocolunifying,nie2026dsgymholisticframeworkevaluating}, and computer use~\citep{lai2025computerrlscalingendtoendonline,osworld, sun2025digidatatrainingevaluatinggeneralpurpose}. However, applying the same approach to spreadsheets is challenging: Unlike many web or software tasks where success can be validated by unit tests or binary completion signals, \emph{final spreadsheets} are produced by a long sequence of operations involving values, formulas, and layout. This leads to significant difficulties: \textcircled{1} collecting sufficient initial–final spreadsheet pairs for training is expensive and difficult to scale for RL training; and \textcircled{2} without step-by-step supervised fine-tuning data, which is even more costly to obtain, the agent must begin RL from a weak interaction policy. This makes a spreadsheet-specific harness essential for providing a structured action space and workflow prior that enable a meaningful initial success rate.

In this paper, we introduce \textbf{Spreadsheet-RL}, a framework for building specialized spreadsheet agents that, to the best of our knowledge, features the first end-to-end RL post-training method for the spreadsheet domain. Spreadsheet-RL differs from prior prompt-driven works by utilizing on-policy RL (e.g., GRPO~\citep{shao2024deepseekmathpushinglimitsmathematical}) in a real-world spreadsheet environment. 
First, for training data collection, \textbf{Spreadsheet-RL} features the automated \textit{Spreadsheet Data Agent} that collects and constructs large-scale realistic spreadsheet tasks for outcome-based rewards across specialized domains such as finance, human resources, and supply chain management; next, for performing operations, the \textit{Spreadsheet Gym}---a multi-turn interactive Microsoft Excel environment integrated with a code sandbox, supports a broad range of advanced Excel functionalities.
Finally, \textbf{Spreadsheet-RL} combines the components into a purpose-built asynchronous RL training framework that interfaces seamlessly with long-horizon, multi-turn spreadsheet interactions, supported by a carefully designed agent harness that incorporates a comprehensive tool set, refined tool-routing rules, and workflow for spreadsheet tasks.

We apply \textbf{Spreadsheet-RL} to the Qwen3 series Large Language Models~\citep{yang2025qwen3technicalreport} with GRPO \citep{Guo_2025} objectives to build specialized spreadsheet agents for evaluation on \textcircled{1} SpreadsheetBench~\citep{spreadsheetbench}, the largest open-source benchmark, and \textcircled{2} \textit{Domain-Spreadsheet}, the first open-source domain-specific spreadsheet benchmark we curate. On SpreadsheetBench, \textbf{Spreadsheet-RL} improves Qwen3-4B-Thinking-2507 from \textbf{12.0\%} to \textbf{23.4\%} Pass@1. These gains, summarized in Figure~\ref{fig:intro}, show how spreadsheet-native harness design, richer tool access, and RL post-training each improve the same 4B open-source base model. Our results also demonstrate that \textbf{Spreadsheet-RL} generalizes across real-world spreadsheet tasks from specialized domains: on Domain-Spreadsheet, \textbf{Spreadsheet-RL} improves overall pass@1 from \textbf{8.4\%} to \textbf{17.2\%} (Table~\ref{tab:domain_spreadsheet_stats_results}). Finally, training dynamics and qualitative analysis show that RL improves not only final accuracy but also interaction efficiency and protocol-following behavior (Figure~\ref{fig:training_dynamics}, Appendix~\ref{sec:appendix_post_rl_cases}).

Overall, \textbf{Spreadsheet-RL} establishes outcome-based RL as a practical and effective post-training paradigm for spreadsheet automation. By releasing the data, environment, harness, training pipeline, and model, Spreadsheet-RL provides an end-to-end reproducible foundation and the first open playground for future research on spreadsheet agents.

\section{Related Work}
This section overviews related work in automating spreadsheet workflows and recent developments and benchmark datasets for spreadsheet workflows necessary for applying RL fine-tuning to the spreadsheet domain.

\paragraph{Spreadsheet Workflow Automation.}
There exists a long line of work in automating spreadsheet manipulation covering a wide variety of techniques. Early work typically targeted specific, well-scoped tasks, such as automated string processing~\citep{string_process}, detecting spreadsheet code smells~\citep{detectsmell}, or clustering related cells~\citep{cell_cluster}. More recent works such as SheetCopilot and SheetAgent~\citep{sheetcopilot, sheetagent} utilize AI agents, formulating the desired spreadsheet operations in natural language while the agents interact with spreadsheets via programmatic interfaces such as Python-based environments or Excel tool APIs (e.g., MCP servers)~\citep{musa_excel_mcp_server}.
These existing agent-based approaches largely focus on inference-time design and prompt engineering; in comparison, \textbf{Spreadsheet-RL} uniquely performs \emph{model-side} agentic training via RL fine-tuning, enabling it to achieve significantly higher performance on more complex, multi-step spreadsheet workflows (\cref{sec:experiments}).
\begin{figure*}[t]
  \centering
  \includegraphics[width=\textwidth]{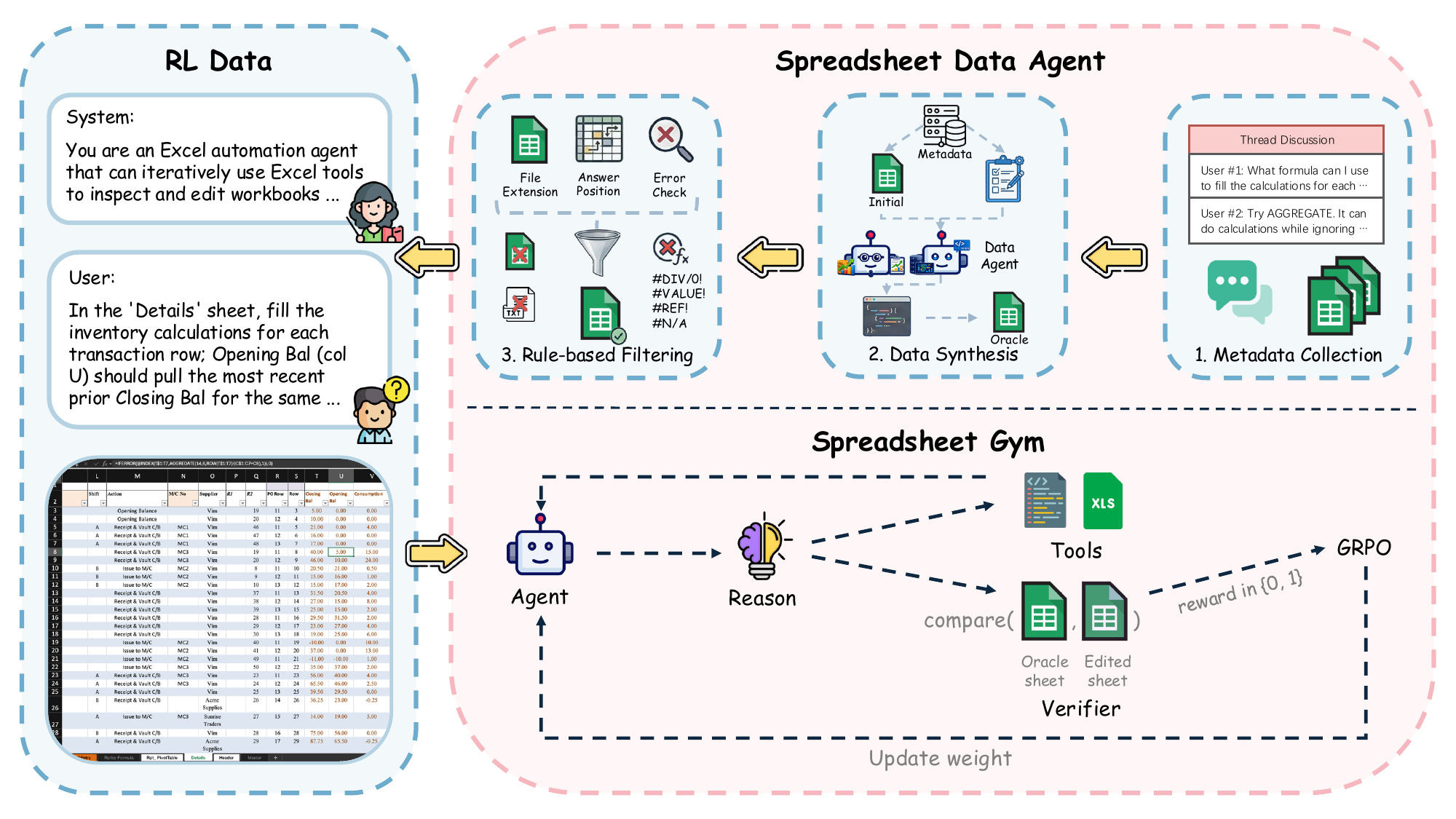}
  \caption{Overview of \textbf{Spreadsheet-RL}. We construct an RL dataset from real spreadsheet problems, consisting of natural-language task descriptions, initial spreadsheets, and oracle final spreadsheets. A policy LLM interacts with \textit{Spreadsheet Gym} (a real Excel environment) to generate multi-step spreadsheet edits through interleaved reasoning and tool use. Rewards are computed by comparing the predicted final spreadsheet against the oracle outcome. The LLM policy is optimized with GRPO.}
  \label{fig:thumbnail}
\end{figure*}

\paragraph{Benchmark Datasets for Spreadsheets.}
Several recent works have introduced benchmark datasets and/or data collection methods for evaluating spreadsheet workflows.
SpreadsheetBench~\citep{spreadsheetbench} collects 912 paired initial–final spreadsheets from online forums with verification by 20 experts.
SheetCopilot~\citep{sheetcopilot} synthesizes tasks from 28 workbooks.
SheetAgent~\citep{sheetagent} performs evaluation with spreadsheet-adjacent, table-centric QA benchmarks such as WikiTableQuestions~\citep{pasupat-liang-2015-compositional} and TabFact~\citep{2019TabFactA}.
OpenAI Agent uses proprietary internal spreadsheet datasets from domains such as investment banking to evaluate its performance~\citep{openaiagent}.
That is, there currently does not exist an open-source framework dedicated to spreadsheets that features automated web-scale data collection; Spreadsheet-RL fills this gap by introducing a fully open-source, agent-driven pipeline for constructing large-scale spreadsheet workflows (i.e., initial--final spreadsheet pairs) from a wide variety of domains for benchmarking, which effectively supports RL training and evaluation of spreadsheet agents.

\section{Spreadsheet-RL}

This section overviews the Spreadsheet-RL framework. We formulate Spreadsheet-RL's task in \cref{sec:methodology_prelimiary}, detail Spreadsheet-RL's automated task construction and interactive spreadsheet agent harness (via the \textit{Spreadsheet Gym}) in \cref{sec:methodology_gym}, present details of Spreadsheet-RL's asynchronous RL training pipeline in \cref{sec:methodology_rl}, and describe a new, open-source dataset, \textit{Domain-Spreadsheet}, which we curate for Spreadsheet-RL's evaluation in \cref{sec:methodology_dataset}.

\subsection{Task Formulation}
\label{sec:methodology_prelimiary}
Spreadsheet-RL follows the task formulation defined in SpreadsheetBench~\citep{spreadsheetbench}, where each task consists of (potentially multiple) \textbf{initial spreadsheets $D_i$}, a \textbf{natural-language instruction $T$}, an \textbf{oracle final spreadsheet $D_O$} (used for RL) representing the correct post-operation result of each task, and the \textbf{manipulation regions} $M$ (e.g., target sheets and cell ranges) for reward computation only. The spreadsheet agent $A$---practically, a large language model that interleaves reasoning with programmatic interactions with the spreadsheet agent harness, must follow $T$ to execute a sequence of spreadsheet operations $T_A$ to arrive at a final spreadsheet $D_A$ that matches the oracle $D_O$.

\subsection{Spreadsheet Data and Environment}
\label{sec:methodology_gym}
To construct the initial dataset and environment for Spreadsheet-RL, we introduce \textit{Spreadsheet Data Agent}, which automates spreadsheet task generation (\cref{sec:methodology_task}), and \textit{Spreadsheet Gym} with agent harness design, which enables LLM agents to interactively execute spreadsheet operations in real Microsoft Excel while interleaving these actions with reasoning traces (\cref{sec:methodology_operations}).

\subsubsection{Task Generation with Spreadsheet Data Agent}
\label{sec:methodology_task}
Large-scale spreadsheet task data is expensive to create from scratch through human annotation. To address this, we propose an automated \textit{spreadsheet data agent} that constructs corpora of paired initial--final spreadsheets. Prior datasets for spreadsheet operations~\citep{spreadsheet_mot, sheetagent} which are typically small in scale for RL training (only up to 912 spreadsheet pairs~\citep{spreadsheetbench}), rely heavily on humans during data creation, and largely focus on LLM-synthesized, spreadsheet-agent table QA tasks~\citep{2019TabFactA}; in comparison, the spreadsheet data agent preserves realistic spreadsheet-specific task distributions via its scalable and automated collection of real-world spreadsheet problems from trusted sources such as online forums. It then transforms these into ready-to-use initial--final spreadsheet pairs through rigorous rule-based filtering and validation, all without the assistance of human experts.

\paragraph{Seed Metadata Collection.} The spreadsheet data agent first curates seed metadata instances from high-quality \textbf{public accessible} online spreadsheet forum \textbf{ExcelForum}. Desirable seeds are forum posts that contain (1) a user-provided initial spreadsheet $D_i$ and a concrete task utilizing a wide range of advanced operations such as complex formulas, formatting, pivot tables, and VBA/macros, and (2) a discussion thread containing potential solutions to the provided task. Seeds are identified using simple heuristics, such as the presence of an attached spreadsheet and multi-turn response chains. For each seed, the information $T$ includes proposed solutions from the discussion thread, intermediate explanations, and follow-up clarifications about the spreadsheet task.

\paragraph{Oracle Construction via Coding Agents.} The oracle $D_O$ for each seed metadata instance is built via strong coding agents (e.g., Claude Code and Codex). The coding agent is prompted with the aforementioned initial workbook $D$ and the collected task instructions and solution discussions $T$, and is instructed to generate an executable sequence of spreadsheet edits to perform the task. The coding agent executes the generated procedure on $D$ in a real Excel environment (described shortly in \cref{sec:methodology_operations}) and records the resulting spreadsheet as the candidate oracle $D_O$. Finally, quality checking is performed by applying rule-based filtering and verification (e.g., removing samples that trigger Excel errors and automatically validating that all values are computable via formulas) and discarding instances that fail verification.

\subsubsection{Spreadsheet Runtime: Gym with Microsoft Excel and Code Sandbox}
\label{sec:methodology_operations}
The \textit{Spreadsheet Gym} is a multi-turn environment enabling interactions with a real spreadsheet instance, coupled with carefully curated spreadsheet-native tool set and an open-source code sandbox~\citep{bytedanceseedfoundationcodeteam2025fullstackbenchevaluatingllms} that allows the agent to execute Python for auxiliary computation to invoke structured APIs for stateful spreadsheet edits.

\paragraph{Execution fidelity.}
Spreadsheet Gym utilizes Microsoft Excel as its spreadsheet instance, which supports a rich set of advanced features and modern functions, including dynamic array formulas such as \texttt{FILTER}, \texttt{UNIQUE}, \texttt{SORT}, \texttt{TAKE}, and \texttt{MAP}, many of which are lacking in alternative engines such as LibreOffice Calc. This wide range of features present in Excel enables Spreadsheet Gym to perform training and evaluation under realistic and complex execution semantics, ensuring alignment between learned agent behavior and real-world spreadsheet workflows.

\paragraph{Compatibility with RL training.}
Spreadsheet Gym features a \emph{per-rollout, filesystem-isolated workspace} for safe parallel execution, ensuring its compatibility with large-scale RL asynchronous training frameworks such as VeRL~\citep{verl}. Each gym instance is assigned a unique workspace identifier, and all relevant spreadsheet artifacts are read from and written to the corresponding workspace. This prevents cross-trajectory file clobbering and data corruption, enabling efficient and scalable batched rollouts crucial to modern RL training. We discuss this design choice in Appendix~\ref{sec:appendix_workspace_isolation}.

\subsubsection{Spreadsheet-Native Tool Harness.}
\label{sec:methodology_prompt}

\paragraph{Harness Overview.}
Spreadsheet-RL includes a spreadsheet-specific agent harness that is crucial for reliable long-horizon spreadsheet interaction. Unlike general-purpose agent prompts, our harness is tailored to the distinctive nature of spreadsheet tasks: the harness defines a clear role for the agent, routes different spreadsheet operations to specialized tools, and enforces safe tool-calling rules that allow parallel read-only inspection while serializing write operations to avoid conflicting workbook mutations. The harness further guides the agent through an inspect, modify, and verify workflow, encouraging it to first identify relevant ranges, then make minimal necessary edits, and finally verify affected values, formulas, and formatting before continuing, as shown in the prompt below:

\begin{tcolorbox}[
  breakable,
  halign=flush left,
  colback=googlegray!5!white,
  colframe=googlegray!75!black,
  title={\textbf{Spreadsheet-Native Tool Harness Prompt (Compressed)}},
  fonttitle=\bfseries,
  fontupper=\ttfamily,
  width=\linewidth,
]
\small
{\color{googlered}\textbf{Role.}} Edit Excel workbooks to satisfy the user's requested end state. The workbook itself is the answer; do not answer conceptually.\\[0.2em]

{\color{googlered}\textbf{Tool router.}} Use {\color{googleblue}\textbf{find\_cells}} for headers/anchors, {\color{googleblue}\textbf{inspect\_range}} for relevant ranges, {\color{googleblue}\textbf{fill\_formula}} for formula-filled targets, {\color{googleblue}\textbf{clear\_range}} for blank cells, {\color{googleblue}\textbf{delete\_rows/delete\_columns}} for structural deletion, {\color{googleblue}\textbf{recalculate\_and\_read}} after custom formula edits, and {\color{googleblue}\textbf{code\_interpreter}} for custom logic or fallback.\\[0.2em]

{\color{googlered}\textbf{Tool calling.}} Concurrent read-only calls are allowed up to 20 per assistant turn. Write-related calls must be issued one at a time and must not be mixed with read-only calls.\\[0.2em]

{\color{googlered}\textbf{Workflow.}} Inspect small relevant workbook ranges $\rightarrow$ plan the smallest necessary edit $\rightarrow$ modify the workbook $\rightarrow$ verify values, formulas, or formatting $\rightarrow$ fix iteratively if needed $\rightarrow$ save \texttt{data.xlsx} before continuing.
\end{tcolorbox}

This harness defines a spreadsheet-native action space that encodes common spreadsheet semantics directly into tools. A purely general code interface is expressive, but it forces the model to re-implement spreadsheet semantics in ad hoc Python. This is brittle for small and medium LLMs: structural edits can be invalidated by index shifts, formula edits require careful reference translation and string escaping, and many tasks require distinguishing blanking cells from deleting rows or columns. By exposing structured tools for these operations, the harness reduces low-level execution failures and provides a stronger initial interaction policy for RL training without SFT warm-up. The full haharness prompt and detailed tool descriptions are provided in Appendix~\ref{app:harness}.

\subsection{Asynchronous RL Pipeline}
\label{sec:methodology_rl}

Spreadsheet-RL's asynchronous RL training pipeline uses GRPO with verifiable outcome-based rewards. To reduce the difficulty of sparse terminal rewards in long-horizon spreadsheet tasks, the rollout prompt encourages verification observations during interaction with Spreadsheet Gym, while the RL objective itself remains outcome-based.

\subsubsection{GRPO with Outcome-based Reward}
\label{sec:methodology_rl_outcome}
Spreadsheet-RL trains and evaluates a policy LLM to solve spreadsheet tasks via multi-turn interaction with \textit{Spreadsheet Gym}. The model receives the initial spreadsheet $D_i$, the natural-language instruction $T$, and harness details such as available tools and the interaction protocol. At each assistant turn, the model produces reasoning and one or more tool calls; tool responses are returned to the model before the next turn. Motivated by ReAct~\citep{yao2023react}, this interleaves reasoning with action, but specializes the action space to spreadsheet-native tools and verification-oriented workbook reads.

Appendix~\ref{sec:appendix_rollout_trace} depicts an accepted rollout where an LLM agent follows the input prompt: it alternates between natural-language reasoning and spreadsheet-native tool use, with \texttt{code\_interpreter} as a fallback for custom logic, to produce a sequence of spreadsheet edits. The trajectory terminates upon task completion (or a step limit) to produce a final spreadsheet $D_{\text{pred}}$, which is evaluated against the oracle $D_o$ to compute the outcome-based reward $\mathcal{R}$ as follows:

\vspace{-2mm}
\begin{equation}
\mathcal{R}(o)=
\begin{cases}
0, & \text{if no valid output},\\[-0.2em]
\mathrm{allcellsmatch}(D_{\mathrm{pred}}, D_o), & \text{otherwise}.
\end{cases}
\label{eq:reward}
\end{equation}

where $\mathrm{allcellsmatch}(\cdot)$ compares the specified manipulation regions $M$ (target sheets and cell ranges) between the LLM agent-produced final spreadsheet $D_{\text{pred}}$ and $D_o$. In practice, this comparison can incorporate value-level matching with numeric tolerances and, when applicable, formula- or structure-level checks, enabling reliable and verifiable outcome supervision for RL.

\paragraph{Asynchronous reward API.}
In \textit{Spreadsheet Gym}, reward computation is not a lightweight in-process function: faithful evaluation requires opening the edited workbook in Microsoft Excel, triggering recalculation, and comparing the recalculated output against the oracle workbook. Since this process can be slow and depends on  Windows and Excel, we propose an asynchronous submit-and-poll verifier that makes \textbf{Excel-based reward computation scalable for RL training.} We provide implementation details in Appendix~\ref{sec:appendix_verifier_api}.

\paragraph{Training Objective.} \textbf{Spreadsheet-RL} aims to improve the LLM policy $\pi_{\theta}$'s spreadsheet manipulation capabilities (i.e., with spreadsheet gym $G$) without incurring large update steps from the reference model $\pi_{\text{ref}}$ (which are appropriately penalized), with a training objective as follows:
\begin{equation}
\label{eq:rl-tooluse}
\max_{\pi_\theta}\;
\mathbb{E}_{[D_i,T]\sim\mathcal{D},\,y\sim\pi_\theta(\cdot\mid D_i,T;G)}
\!\left[\mathcal{R}(D_i,T,D_o)\right]
-\beta\,\mathbb{D}_{\mathrm{KL}}\!\left[
\pi_\theta(\cdot\mid D_i,T;G)
\,\|\,
\pi_{\mathrm{ref}}(\cdot\mid D_i,T;G)
\right].
\end{equation}
\noindent where $\pi_{\text{ref}}$ is a frozen reference model, $\mathcal{R}$ is the outcome reward, and $\beta>0$ controls the KL Divergence penalty. The pair $[D_i, T]$ denotes a task sampled from $\mathcal{D}$, consisting of an initial spreadsheet $D_i$ and a natural-language instruction $T$. We explicitly condition on $G$ to emphasize that the policy's token generation is interleaved with multi-turn interaction with the spreadsheet environment, and the resulting trajectory determines $D_{\text{pred}}$, and hence the final outcome reward.

Specifically, \textbf{Spreadsheet-RL}'s optimization for parameters $\theta$ builds on GRPO~\citep{Guo_2025}, which estimates baselines from a group of Monte-Carlo sampled rollouts, eliminating the need for a critic, reducing training overhead and costs, and demonstrating strong empirical performance. This efficiency makes GRPO particularly well-suited for Spreadsheet-RL's complex, multi-turn setting where training costs may otherwise be prohibitive. Training objective is in Appendix \cref{eq:grpo_condensed_fig}.

\section{Domain-Spreadsheet Benchmark for Generalization Evaluation}
\label{sec:methodology_dataset}
Spreadsheet-RL contains an accompanying dataset benchmark, \textit{Domain-Spreadsheet}, a domain-specific evaluation set of \textbf{1,660} spreadsheet tasks spanning finance (beginner/intermediate/advanced), supply chain, human resources, sales, and real estate (\cref{tab:domain_spreadsheet_stats_results}). Compared to existing open-source spreadsheet benchmarks which focus on operation-centric tasks, Domain-Spreadsheet emphasizes domain knowledge and professional analytical workflows.

\begin{wrapfigure}{r}{0.46\columnwidth}

\centering
\vspace{-2em}
\includegraphics[width=0.44\columnwidth]{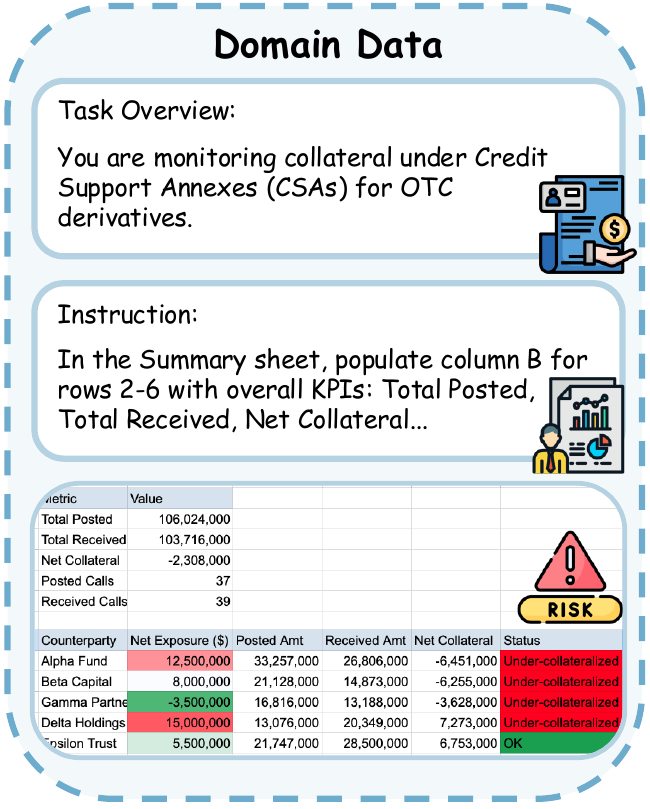}

\caption{Domain-Spreadsheet example: finance spreadsheet data for risk tasks.}
\label{fig:domain}
\vspace{-1em}

\end{wrapfigure}
\paragraph{Domain Metadata Collection.}
Domain-specific knowledge of the aforementioned topics \textbf{is not readily available} in public spreadsheet forums and often requires substantial expert effort to manually annotate~\citep{openai2025introducinggpt52}. Accordingly, Spreadsheet-RL collects Domain-Spreadsheet by (1) first curating domain concepts and professional templates from sources spanning knowledge areas covered by mainstream professional certifications (e.g., CPA, CFA, and FRM for finance, CPIM for supply chain, SHRM, CCP for human resources, and CCIM for real estate) and common-practice analytical workflows in topics including investment banking, asset management, inventory analysis, compensation benchmarking, and property valuation, then (2) prompting the data agent to summarize the collected data into task specifications, which are next passed to the seed metadata construction pipeline to generate the corresponding executable tasks (\cref{sec:methodology_task}). We show one finance data example in Figure~\ref{fig:domain}.

We observe that the collected tasks in Domain-Spreadsheet reflect realistic professional workflows: using finance as an example, building comparable-company analyses with trading multiples, computing Value-at-Risk, and modeling debt-service coverage ratios, and so on. We use Domain-Spreadsheet as one of our benchmark datasets for Spreadsheet-RL's evaluation (\cref{sec:experiments}), to claim generalizability of RL to tasks in different domains.

\section{Spreadsheet-RL Evaluation}
\label{sec:experiments}
This section empirically studies the effectiveness of the Spreadsheet-RL framework.
\subsection{Experiment Setup}
\paragraph{Training Dataset.}
Spreadsheet-RL, and other applicable baselines, are trained using datasets collected by the spreadsheet data agent (\cref{sec:methodology_task}). Specifically, we collect posts \textbf{from after 01/01/2024}, from publiclyly accessible ExcelForum, and collect 18,855 raw discussion threads with a total of 32,691 spreadsheet attachments and 144,694 user replies (i.e., average 7.67 replies per thread), which, after being passed through our data agent and rule-based filtering, results in a training dataset containing \textbf{5928} high-quality tasks, each consisting of a task instruction, initial spreadsheets (2417 tasks have more than 1 initial spreadsheet, shown in \cref{fig:train-sheet-proportion-distribution}), and an oracle final spreadsheet. We summarize the distribution of spreadsheet operations in our training dataset in \cref{fig:verb-noun-phrase-distribution}, and the more detailed row, column, and worksheet count distributions of initial spreadsheets in Appendix~\cref{fig:train_data_stats}.
\begin{table}[t]
\centering
\footnotesize
\setlength{\tabcolsep}{6pt}
\newcolumntype{R}[1]{>{\raggedleft\arraybackslash}p{#1}}
\begin{tabular}{p{6.0cm} R{2.5cm} r r}
\toprule
\textbf{Model} & \textbf{Environment Access} & \textbf{Accuracy Score} & \textbf{Eval. Reference} \\
\midrule
\rowcolor{LightGray}
\multicolumn{4}{c}{\textbf{Closed-source models}}\\
\midrule
GPT-4o \citep{openai2024gpt4ocard}  & OSX, LibreOffice & 16.8 & OpenAI \citep{openaiagent}  \\
GPT-4o \citep{openai2024gpt4ocard} & Windows, Excel & 18.4 & OpenAI \citep{openaiagent} \\

OpenAI o3 \citep{openaio3} & OSX, LibreOffice & 23.3 & OpenAI \citep{openaiagent} \\
ChatGPT agent \citep{openaiagent} & OSX, LibreOffice & 35.3 & OpenAI \citep{openaiagent} \\
Claude Files Opus 4.1 \citep{opus41} & Windows, Excel & 42.9 & Microsoft \citep{microsoft}\\
ChatGPT agent with .xlsx access \citep{openaiagent} & OSX, LibreOffice & 45.5 & OpenAI \citep{openaiagent} \\

Copilot Agent Mode \citep{microsoft} & Windows, Excel & 57.7 & Microsoft \citep{microsoft} \\

\midrule
\rowcolor{LightGray}
\multicolumn{4}{c}{\textbf{Open-source models}}\\
\midrule

Qwen3-4B-Instruct-2507$^{\dagger}$ & Spreadsheet Gym & 9.3 & Ours \\

Qwen3-4B \citep{yang2025qwen3technicalreport} & Spreadsheet Gym & 11.0 & Ours \\

Qwen3-14B \citep{yang2025qwen3technicalreport} & Spreadsheet Gym & 15.0 & Ours \\

Qwen3-8B \citep{yang2025qwen3technicalreport} & Spreadsheet Gym & 15.9 & Ours \\

Qwen3-32B \citep{yang2025qwen3technicalreport} & Spreadsheet Gym & 17.6 & Ours \\

\midrule
\rowcolor{LightGray}
\multicolumn{4}{c}{\textbf{Qwen3-4B-Thinking-2507: From raw base model to Spreadsheet-RL-trained agent}}\\
\midrule
Qwen3-4B-Thinking-2507 & Spreadsheet Gym & 12.0 & Ours \\
+ Spreadsheet-native interaction harness & Spreadsheet Gym & 15.6 & Ours \\
+ Comprehensive spreadsheet-tool access & Spreadsheet Gym & 19.3 & Ours \\
+ Spreadsheet-RL post-training & Spreadsheet Gym & \textbf{23.4} & Ours \\

\bottomrule
\end{tabular}
\vspace{0.2em}
\caption{\textbf{Main Pass@1 results on SpreadsheetBench \citep{spreadsheetbench}}. Spreadsheet-RL improves the accuracy of the open-source Qwen3-4B-Thinking-2507 model from \textbf{12.0} to \textbf{23.4}, with gains from spreadsheet-native interaction design, comprehensive tool access, and RL post-training. $^{\dagger}$ denotes the latest released Qwen3 model, which is not included in the technical report \citep{yang2025qwen3technicalreport}.}
\vspace{-3em}
\label{tab:spreadsheet_bench}
\end{table}

\begin{wraptable}{r}{0.5\textwidth}
\vspace{-1.2em}
\centering
\scriptsize
\setlength{\tabcolsep}{2.6pt}
\renewcommand{\arraystretch}{1.05}
\resizebox{\linewidth}{!}{
\begin{tabular}{@{}lccc|cc@{}}
\toprule
\textbf{Domain} 
& \textbf{\#Eval.} 
& \textbf{Sheets} 
& \textbf{Input} 
& \multicolumn{2}{c}{\textbf{Pass@1}} \\
\cmidrule(lr){5-6}
& 
& \textbf{Mean} 
& \textbf{Med. KB} 
& \textbf{Base} 
& \textbf{RL} \\
\midrule
Finance-B    & 597 & 2.0 & 9.5  & 15.6 & \textbf{29.3} \\
Finance-I    & 388 & 2.6 & 9.9  & 7.7 & \textbf{16.2} \\
Finance-A    & 135 & 3.0 & 10.4 & 8.1 & \textbf{19.3} \\
Supply Chain & 180 & 3.4 & 16.0 & 1.1 & \textbf{5.0} \\
HR           & 185 & 3.3 & 13.7 & 0.5 & \textbf{3.2} \\
Sales        & 86  & 3.1 & 14.1 & 1.2 & \textbf{5.8} \\
Real Estate  & 89  & 3.5 & 12.8 & \textbf{1.1} & \textbf{1.1} \\
\midrule
\textbf{Overall} 
& \textbf{1,660} 
& \textbf{2.7} 
& \textbf{10.3} 
& 8.4 
& \textbf{17.2} \\
\bottomrule
\end{tabular}
}
\vspace{0.3em}
\caption{\textbf{Domain-Spreadsheet statistics and results.} We report evaluation-rollout counts, workbook statistics, and Qwen3-4B-Thinking-2507's Pass@1 before and after Spreadsheet-RL training.}
\label{tab:domain_spreadsheet_stats_results}
\vspace{-4.0em}
\end{wraptable}

\paragraph{Training Config.} Unless otherwise specified, we fine-tune Qwen3-4B-Thinking-2507~\citep{yang2025qwen3technicalreport} with Spreadsheet-RL for 60 training steps. We choose Qwen3-4B-Thinking-2507 as our base model because, among the non-trained Qwen3 variants evaluated on SpreadsheetBench (Table~\ref{tab:spreadsheet_bench}), it offers a strong accuracy--cost trade-off at the 4B scale. RL Training is conducted in our \textit{Spreadsheet Gym} (\cref{sec:methodology_gym}) which is implemented via the agent loop in VeRL framework~\citep{verl} to support multi-turn asynchronous RL, with Microsoft Excel 365 (version 2512) as the execution environment. Additional implementation and training hyperparameter details are provided in \ref{sec:appendix_hparams}.

\paragraph{Evaluation Datasets.} 
\begin{itemize}[leftmargin=*, nosep]
    \item \textbf{SpreadsheetBench}~\citep{spreadsheetbench} contains \textbf{\textbf{912}} unique tasks, each with an initial--final spreadsheet pair, and each task is further instantiated into three distinct but highly similar test cases. We strictly follow this dataset's evaluation protocol, using our Excel-based environment and the same LLM decoding settings as in training rollouts. 
    \item \textbf{Domain-Spreadsheet} (\cref{sec:methodology_dataset}) contains \textbf{1,660} unique tasks spanning finance, supply chain management, human resources, sales, and real estate, enabling evaluation of cross-domain generalization (\cref{tab:domain_spreadsheet_stats_results}).
\end{itemize}

\paragraph{Evaluation Metrics.}
Our primary evaluation metric is accuracy under exact spreadsheet manipulation–region success, i.e., $\mathrm{allcellsmatch}(D_{\text{pred}}, D_o)$ (Eq.~\ref{eq:reward}), for which we report Pass@1. We compare with a small tolerance threshold for numerical cells, use exact matching for text cells, and compare canonicalized formula strings and/or evaluated values (reporting both when applicable) for formula cells. Appendix~\ref{sec:appendix_sheetagent_baseline} discusses why we do not include a direct SheetAgent rerun as a quantitative baseline.

\subsection{Main Results}
Table~\ref{tab:spreadsheet_bench} reports Pass@1 on SpreadsheetBench. Starting from the raw Qwen3-4B-Thinking-2507 base model at 12.0\%, Spreadsheet-native interaction harnessing raises accuracy to 15.6\%, comprehensive spreadsheet-tool access further raises it to 19.3\%, and Spreadsheet-RL post-training reaches 23.4\%. This staged improvement demonstrates that outcome-based RL fine-tuning is most effective when combined with scalable task construction (\cref{sec:methodology_task}), a faithful Excel runtime, and the inspect-modify-verify interaction protocol embodied by the harness (\cref{sec:methodology_prompt}; Appendix~\ref{sec:appendix_rollout_trace}). Compared to representative proprietary spreadsheet-agent results reported in prior work, our RL-trained 4B open-source spreadsheet agent surpasses the OpenAI o3 baseline while operating entirely within our reproducible \textit{Spreadsheet Gym} environment.

\paragraph{Generalizability of RL Fine-Tuning.}
We study whether a spreadsheet agent trained with Spreadsheet-RL generalizes well to real-world, domain-specific spreadsheet tasks in \textbf{Domain-Spreadsheet}, despite being trained solely on operation-focused data curated from discussion forums. Table~\ref{tab:domain_spreadsheet_stats_results} reports pass@1 over 1,660 evaluation rollouts from the Domain-Spreadsheet logs. Spreadsheet-RL improves overall pass@1 from 8.4\% to 17.2\%, with the largest gains on finance workflows. Gains also appear in supply chain, HR, and sales; real estate remains unchanged at 1.1\%, indicating that this slice remains challenging for the current 4B spreadsheet agent.

\begin{figure}[t]
  \centering
  \includegraphics[width=0.98\linewidth]{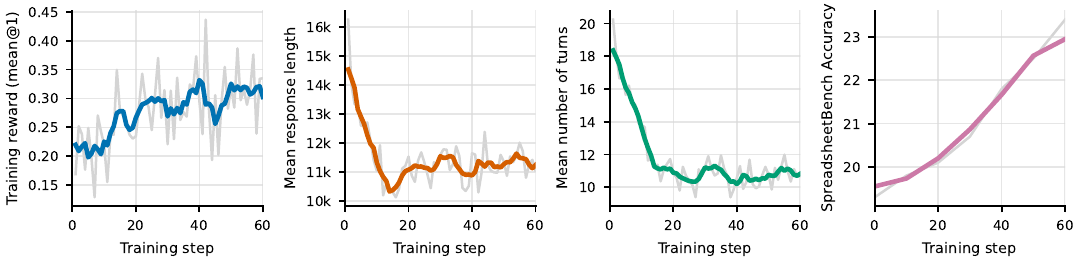}
  \caption{\textbf{RL training dynamics for Qwen3-4B-Thinking-2507.} All panels are constructed from training dynamics during RL training, including mean training reward, mean response length, number of mean turns, and accuracy ccuracy is measured every ten steps.}
  \label{fig:training_dynamics}
  \vspace{-1.5em}
\end{figure}

\paragraph{RL Training Dynamics.}
We track RL training dynamics throughout fine-tuning (Appendix~\ref{fig:training_dynamics}). The smoothed training reward rises from roughly 0.21 in the first steps to 0.33 by step 60, while SpreadsheetBench accuracy improves from 19.3\% at step 0 to 23.4\% at step 60. The same run records show that interaction cost becomes more controlled: mean response length drops from approximately 16k near the start to approximately 11k by step 60, and mean interaction interaction turns fall from roughly 20 to about 11. These dynamics support the main finding that RL training improves both final task success and the efficiency of workbook-editing rollouts.

\subsection{Pre-RL Tool Interface and Post-RL Ablation.}
The staged Qwen3-4B-Thinking-2507 block in Table~\ref{tab:spreadsheet_bench} isolates the effect of Spreadsheet Gym's tool interface before RL post-training. These pre-RL rows are not the final Spreadsheet-RL trained result and should be read as tool-interface evidence: the spreadsheet-native interaction harness raises Pass@1 from 12.0\% to 15.6\%, and comprehensive spreadsheet-tool access raises it further to 19.3\%. The minimal tool setting contains only \texttt{code\_interpreter} and \texttt{recalculate\_and\_read}; the comprehensive tool setting additionally includes \texttt{inspect\_range}, \texttt{find\_cells}, \texttt{fill\_formula}, \texttt{clear\_range}, \texttt{delete\_rows}, and \texttt{delete\_columns}. These gains indicate that spreadsheet-native tools are useful even before RL fine-tuning, because they remove low-level failure modes that are not central to the stasks. We provide representative pre-RL failure cases in Appendix~\ref{sec:appendix_pre_rl_cases} and post-RL behavior comparisons in Appendix~\ref{sec:appendix_post_rl_cases}.

\section{Conclusion}
\label{sec:conclusion}

We present Spreadsheet-RL, the first end-to-end RL method specifically designed for training spreadsheet agents. In contrast to prior prompt-driven spreadsheet agents that depend on general-use and often proprietary LLM advances, Spreadsheet-RL establishes outcome-based, on-policy RL as a core mechanism for learning long-horizon spreadsheet workflows directly through environment interaction.
Spreadsheet-RL unifies large-scale, domain-specific spreadsheet task construction with a multi-turn interactive Excel gym environment. On SpreadsheetBench, staged harness and RL improvements raise Pass@1 for a 4B open-source model from \textbf{12.0\%} to \textbf{23.4\%}. We also evaluate on our domain-specific benchmark, Domain-Spreadsheet, to study cross-domain generalization, where Spreadsheet-RL improves overall pass@1 from \textbf{8.4\%} to \textbf{17.2\%} (Table~\ref{tab:domain_spreadsheet_stats_results}). Our results establish RL fine-tuning as a promising paradigm for scalable and reproducible spreadsheet automation. We discuss the limitations and broader impact of Spreadsheet-RL in Appendix~\ref{sec:limitations} and~\ref{sec:impact}, respectively.

\section*{Acknowledgements}
We would like to acknowledge Zhiqing Sun, Xiaoqi Ren, and Chengxiang Zhai for their valuable suggestions and contributions to our Spreadsheet RL project.

Mingyuan, Banghao, Klara and Minjia were supported by the National Science Foundation grants NSF 2106592,  NSF 1900875, and NSF 2441601. Any results and opinions are our own and do not represent views of National Science Foundation.

UIUC researchers used the Delta advanced computing and data resource which is supported by the National Science Foundation (award OAC 2005572) and the State of Illinois. Delta is a joint effort of the University of Illinois Urbana-Champaign and its National Center for Supercomputing Applications.

\newpage
\bibliographystyle{abbrvnat}
\bibliography{neurips_2026}
\appendix

\newpage
\appendix
\onecolumn
\section{Appendix}
\label{sec:appendix}
\subsection{Limitations}
\label{sec:limitations}
Spreadsheet-RL provides an open research foundation for studying RL post-training in spreadsheet-based data workflows. However, due to resource constraints, our current experiments focus on relatively lightweight open-source models, and we do not report training results for larger dense models or mixture-of-experts (MoE) models. We leave scaling Spreadsheet-RL to larger model families as an important direction for future work, and hope that the release of our data, environment, harness, and training pipeline will make such exploration more accessible to the community.

\subsection{Why SheetAgent Is Not Included as a Direct Baseline}
\label{sec:appendix_sheetagent_baseline}
SheetAgent~\citep{sheetagent} is an important prior spreadsheet-agent design, but we do not include a direct SheetAgent rerun as a quantitative baseline in Table~\ref{tab:spreadsheet_bench}. The system is tightly coupled to a particular model backend, prompt stack, code-retrieval component, and spreadsheet execution setting: its Planner, Informer, and Retriever interact through model-generated code and retrieval from a configured code corpus, so changes in the LLM API, spreadsheet runtime, or dataset interface can substantially affect the final workbook edits. In our reproduction attempt, the public release did not provide a complete turnkey evaluation path for all model backends and spreadsheet environments considered in our study; for example, the Retriever depends on an external Milvus setup, and the released data/evaluation artifacts cover only a subset of the original SheetRM setting. When adapted directly to our SpreadsheetBench evaluation protocol and spreadsheet environment, the system frequently failed before producing valid spreadsheet edits or produced outputs incompatible with our verifier, yielding near-zero Pass@1 in pilot runs. We therefore exclude this rerun from quantitative baselines, since it would measure an incomplete cross-environment reproduction rather than the SheetAgent method itself. Instead, we cite SheetAgent as prior work and compare against baselines whose SpreadsheetBench numbers are reported by prior work or whose runs are reproducible in our Spreadsheet Gym.

\subsection{Broader Impact}
\label{sec:impact}

Spreadsheet-RL aims to make spreadsheet automation more accessible, reproducible, and reliable. By training open-source agents to operate in realistic spreadsheet environments, our framework can help users automate repetitive data-centric workflows in areas such as finance, supply chain, human resources, sales, and personal productivity. The release of our data, environment, harness, training pipeline, and model may also support future research on open, verifiable agents for productivity software.

At the same time, spreadsheet agents may introduce risks when used in high-stakes settings. Incorrect formulas, unintended structural edits, or subtle formatting errors can affect downstream decisions, especially if users over-trust automated outputs. Publicly collected or synthesized training data may also contain biases or domain gaps. Therefore, we view Spreadsheet-RL as a research foundation rather than a fully deployable decision-making system. Practical deployment should include human review, transparent edit logs, stronger verification tools, privacy safeguards, and domain-specific safety checks.

\subsection{Complete Spreadsheet-Native Tool Harness Prompt}
\label{sec:appendix_rollout_prompt}
\begin{tcolorbox}[
    breakable,
    halign=flush left,
    colback=googlegray!5!white,
    colframe=googlegray!75!black,
    title={\textbf{Spreadsheet-Native Tool Harness Prompt}},
    fonttitle=\bfseries,
    fontupper=\ttfamily,
    width=\linewidth,
]
\small
{\color{googlered}\textbf{Role.}} You are an AI assistant that edits Excel workbooks to satisfy the user's requested end state. The workbook itself is the answer. Do not answer conceptually.\\[0.2em]

{\color{googlered}\textbf{Tool router.}}\\
{\color{googleyellow}\textbf{- Inspection:}} Use {\color{googleblue}\textbf{find\_cells}} to locate headers, anchors, or text in a sheet. Use {\color{googleblue}\textbf{inspect\_range}} to inspect small relevant ranges.\\
{\color{googleyellow}\textbf{- Edits:}} Use {\color{googleblue}\textbf{fill\_formula}} when target cells should contain formulas. Use {\color{googleblue}\textbf{clear\_range}} when the desired end state is blank cells. Use {\color{googleblue}\textbf{delete\_rows/delete\_columns}} when the workbook should physically lose rows or columns.\\
{\color{googleyellow}\textbf{- Verification and fallback:}} Use {\color{googleblue}\textbf{recalculate\_and\_read}} after custom formula edits to recalculate in Excel and read back specified ranges. Use {\color{googleblue}\textbf{code\_interpreter}} for custom inspection or edits not covered by specialized tools, or when specialized tools fail.\\[0.2em]

{\color{googlered}\textbf{Tool calling.}} For read-only calls, output multiple concurrent tool calls in one assistant turn when useful, up to 20 total calls. For write-related calls, output exactly one tool call in that assistant turn, and do not mix read-only and write-related calls.\\[0.2em]

{\color{googlered}\textbf{Workflow.}} Inspect small relevant ranges on the workbook to understand context. Make a plan and then make the smallest necessary changes. Verify before stopping: for formula/value tasks, inspect the edited cells; for structure/formatting tasks, inspect the relevant metadata. If verification fails, identify the issue and fix it iteratively. Always operate on {\color{googleblue}\textbf{data.xlsx}}, the persistent working workbook.
\end{tcolorbox}

\subsection{Inspect-Modify-Verify Example Rollout}
\label{sec:appendix_rollout_trace}
\begin{tcolorbox}[
  breakable,
  halign=flush left,
  colback=googlegray!5!white,
  colframe=googlegray!75!black,
  title={\textbf{Inspect-Modify-Verify Example Rollout}},
  fonttitle=\bfseries,
  fontupper=\ttfamily,
  width=\linewidth,
]
\small

{\color{googleyellow}\textbf{User:}} Import a value from \texttt{Sheet2} to \texttt{Sheet1}: if \texttt{Sheet1!B1} and \texttt{Sheet2!B1} match, fill the blank target cell \texttt{Sheet1!D1} with \texttt{Sheet2!D1}.\\[0.25em]

{\color{googleblue}\textbf{Assistant:}} {\color{googlegray}\textbf{<think>}} First verify that the matching key in \texttt{B1} is identical across both sheets. {\color{googlegray}\textbf{</think>}}
{\color{googlegray}\textbf{<tool\_call>}} \emph{(\texttt{inspect\_range} on \texttt{Sheet1!B1} and \texttt{Sheet2!B1})} {\color{googlegray}\textbf{</tool\_call>}}\\[0.25em]

{\color{googlegreen}\textbf{Tool:}} {\color{googlegray}\textbf{<tool\_response>}} Both cells contain \texttt{124 bent street}. {\color{googlegray}\textbf{</tool\_response>}}\\[0.25em]

{\color{googleblue}\textbf{Assistant:}} {\color{googlegray}\textbf{<think>}} Since the keys match, inspect the source value in \texttt{Sheet2!D1}. {\color{googlegray}\textbf{</think>}}
{\color{googlegray}\textbf{<tool\_call>}} \emph{(\texttt{inspect\_range} on \texttt{Sheet2!D1})} {\color{googlegray}\textbf{</tool\_call>}}\\[0.25em]

{\color{googlegreen}\textbf{Tool:}} {\color{googlegray}\textbf{<tool\_response>}} \texttt{Sheet2!D1} has value \texttt{15100}. {\color{googlegray}\textbf{</tool\_response>}}\\[0.25em]

{\color{googleblue}\textbf{Assistant:}} {\color{googlegray}\textbf{<think>}} Copy the inspected value into the blank target cell and save \texttt{data.xlsx}. {\color{googlegray}\textbf{</think>}}
{\color{googlegray}\textbf{<tool\_call>}} \emph{(\texttt{code\_interpreter}: load workbook, set \texttt{Sheet1!D1 = Sheet2!D1}, save, and open the workbook again to check the written value)} {\color{googlegray}\textbf{</tool\_call>}}\\[0.25em]

{\color{googlegreen}\textbf{Tool:}} {\color{googlegray}\textbf{<tool\_response>}} Code executed successfully \emph{(Shows the two cells have the same value)} . {\color{googlegray}\textbf{</tool\_response>}}\\[0.25em]

{\color{googleblue}\textbf{Assistant:}} The task has been completed successfully. Data from \texttt{Sheet2!D1} has been written into \texttt{Sheet1!D1} after confirming that the \texttt{B1} values match.
\end{tcolorbox}

\subsection{Qualitative Case Studies: Pre-RL Tool Interface Failures}
\label{sec:appendix_pre_rl_cases}
We examined failed evaluation rollouts from the minimal tool setting to understand common failure modes when \texttt{code\_interpreter} is responsible for workbook edits and \texttt{recalculate\_and\_read} is used only for recalc/readback. These examples do not imply that Python cannot express the desired operations; rather, they show how model-generated code can mis-handle low-level spreadsheet semantics that structured tools make explicit. Two representative failures motivate the structured edit tools in \cref{sec:methodology_prompt}.

\begin{tcolorbox}[
  breakable,
  halign=flush left,
  colback=googlegray!5!white,
  colframe=googlegray!75!black,
  title={\textbf{Minimal-Interface Failure Modes in Evaluation Rollouts}},
  fonttitle=\bfseries,
  fontupper=\ttfamily,
  width=\linewidth,
]
\small
{\textbf{Case 1: Header search with structural deletion.}} A task asks the agent to delete every column whose row-1 header contains \texttt{/description}. In a minimal-interface rollout, the generated code resembles \texttt{for col in range(1, sheet.max\_column + 1): ... sheet.delete\_cols(col)}, deleting immediately after each match. Because deleting a column shifts all subsequent columns left, later indices no longer refer to the originally inspected columns, causing the rollout to skip or delete the wrong columns. The \texttt{delete\_columns} tool instead exposes structural deletion as a single spreadsheet operation after the target ranges are identified.\\[0.4em]

{\textbf{Case 2: Formula propagation by string templating.}} A task asks the agent to modify a VLOOKUP formula in \texttt{G3:G58}. In a minimal-interface rollout, the model hand-constructs formula strings inside a Python loop, including an \texttt{IFERROR(VLOOKUP(...), "")} branch with embedded quotation marks, and fails to escape the string correctly while also taking responsibility for row-specific references. The \texttt{fill\_formula} tool shifts this burden to the harness: the model provides the target range and the top-left formula template, and the tool translates relative references and verifies recalculation before overwriting the workbook.
\end{tcolorbox}

\subsection{Qualitative Case Studies: Post-RL Rollout Behavior}
\label{sec:appendix_post_rl_cases}
To better understand how Spreadsheet-RL changes rollout behavior beyond final-task accuracy, we compare two checkpoints of the same 4B agent: an early checkpoint at step~0 and a later checkpoint at step~50. Both corresponding rollout logs contain 2,726 SpreadsheetBench rollouts, but the step~50 checkpoint is more concise (mean assistant-output length 38,965 vs.\ 51,732 characters). Using simple string matching on assistant outputs, the step~50 checkpoint more often states an explicit \texttt{Alternative plan} when backtracking (4.3\% vs.\ 1.2\%). Conversely, the step~0 checkpoint more frequently relies on speculative validation language (e.g., ``should work'', 60.6\% vs.\ 55.1\%) and more often admits being stuck (``I'm really stuck'', 21.3\% vs.\ 0.8\%). The excerpts below illustrate these qualitative differences.

\begin{tcolorbox}[
  breakable,
  halign=flush left,
  colback=googlegray!5!white,
  colframe=googlegray!75!black,
  title={\textbf{Rollout Excerpts (Step~50 vs.\ Step~0)}},
  fonttitle=\bfseries,
  fontupper=\ttfamily,
  width=\linewidth,
]
\small
{\textbf{Case 1: Alternative plan vs.\ speculative ``should work'' (Task 1\_536-37).}}\\
\textcolor{googleblue}{\textbf{Step~50:}} ``Alternative plan: Since the user has the 'NO Listing' in their row 4, we can find the exact cell with 'NO Listing' and use that.''\\
\textcolor{googlered}{\textbf{Step~0:}} ``The code should work.''\\[0.4em]

{\textbf{Case 2: Stable debugging vs.\ crash-and-recover (Task 1\_178-22).}}\\
\textcolor{googleblue}{\textbf{Step~50:}} ``So, it's possible that there are no rows matching the criteria.''\\
\textcolor{googlered}{\textbf{Step~0:}} ``The previous attempt to create 'Sheet2' failed with a KeyError on index 1.''\\[0.4em]

{\textbf{Case 3: Prompt scaffold and protocol compliance (Task 1\_39180).}}\\
\textcolor{googleblue}{\textbf{Step~50:}} ``Plan: Check the structure of the Excel file, confirm that column C has data starting from row 3 to 57.''\\
\textcolor{googlered}{\textbf{Step~0:}} ``Okay, the user is saying there's no new tool execution since the last submission, so I need to make changes with code\_interpreter.''
\end{tcolorbox}

\subsection{Harness Details}
\label{app:harness}
\paragraph{Spreadsheet-Native Tool Interface.}
A purely general code interface is expressive, but it forces the model to re-implement spreadsheet semantics in ad hoc Python. This is brittle for small and medium LLMs: structural edits can be invalidated by index shifts, formula edits require careful reference translation and string escaping, and many tasks require distinguishing blanking cells from deleting rows or columns. To address these issues, Spreadsheet Gym exposes a structured spreadsheet-native tool interface that covers common spreadsheet operations while retaining \texttt{code\_interpreter} for custom logic and fallback.

\paragraph{Workbook inspection.}
\texttt{find\_cells} locates headers, anchors, and textual cues in a worksheet, while \texttt{inspect\_range} reads targeted A1 ranges and can optionally include formulas, formatting, merged-cell metadata, validations, and other local sheet details. These tools encourage the model to inspect small relevant regions before editing, rather than loading the entire workbook or hallucinating spreadsheet state.

\paragraph{Spreadsheet-native edits.}
\texttt{fill\_formula} fills a formula over a rectangular range from a top-left formula template, translating relative row and column references across the target range and recalculating before writeback. This addresses a recurring failure mode in which models hand-write repeated formulas but fail to adjust references correctly across rows or columns. \texttt{clear\_range} empties cell contents while preserving workbook structure, which is necessary because models often conflate clearing cells with deleting rows or columns. \texttt{delete\_rows} and \texttt{delete\_columns} perform structural deletion with the expected shift-up or shift-left semantics, avoiding common code-based mistakes such as deleting columns while iterating left-to-right through changing indices.

\paragraph{Verification and fallback.}
\texttt{recalculate\_and\_read} recalculates the workbook in Excel and reads requested ranges from the recalculated file, allowing the model to verify custom formula edits under faithful spreadsheet semantics. \texttt{code\_interpreter} remains available for custom inspection, data transformation, formatting, and operations not covered by specialized tools, as well as for fallback when a structured tool cannot express the desired edit.

\subsection{RL-Training Compatibility and Workspace Isolation}
\label{sec:appendix_workspace_isolation}

\paragraph{Design motivation.}
LLM RL environments are commonly deployed in two broad ways. Service-style environments run as separate HTTP servers (often packaged with Docker or hosted services), which can scale independently from the trainer but add network hops, request serialization, session management, and deployment overhead. In-process environments avoid this communication cost by running inside the trainer's Python environment, but the environment state is then coupled to the trainer process and can easily become non-isolated unless the implementation adds its own state-management layer. Spreadsheet Gym uses a third design point for spreadsheet state: the interaction loop and most spreadsheet tools remain lightweight from the trainer's perspective, while each rollout receives a dedicated filesystem workspace for mutable workbook artifacts.

\paragraph{Per-rollout filesystem state.}
For each rollout trajectory, the agent loop generates a unique workspace identifier and seeds a workspace. The initial workbook is copied into that workspace as the model-facing \texttt{data.xlsx}; subsequent tool calls resolve relative workbook paths inside the same directory and write modified artifacts there. The reward path also reads from the workspace while the sandbox mutates the seeded workbook in place. This layout makes the workbook state explicit, local, and addressable across the whole trajectory.

\paragraph{Alternatives considered.}
\textbf{Separate environment server per rollout.} Spreadsheet rollouts contain many small operations, such as inspecting a range, locating headers, clearing cells, filling formulas, or deleting rows/columns. Forcing each of these stateful spreadsheet operations through a separate environment server would add avoidable request overhead and require additional lifecycle management for thousands of concurrent rollout sessions. Instead, Spreadsheet Gym isolates the mutable workbook state directly in the filesystem and reserves external services for the components that genuinely require them: sandboxed code execution and faithful Excel-based recalculation/reward computation. This preserves the scalability benefits of batched asynchronous rollouts without requiring a heavyweight environment process for every trajectory.
\textbf{Shared in-process state.}
A non-isolated in-process design would be simpler, but it is unsafe for on-policy RL. Modern training batches sample many trajectories for the same or related tasks, and these trajectories may execute tools concurrently. If they share a workbook path, one rollout can overwrite another rollout's intermediate edits, leak state across samples, or corrupt the final file used for reward computation. Per-rollout workspaces remove this failure mode: every trajectory owns its workbook files, and tools reject missing or invalid workspace identifiers rather than silently falling back to a shared dataset directory.

\paragraph{Concurrency and cleanup.}
Workspace preparation is guarded by lock files and a manifest so concurrent requests for the same workspace cannot race during seed-copy initialization. Tools resolve files only under the expected workspace root. After reward submission, the workspace and associated lock artifacts can be removed automatically, which keeps long-running RL jobs from accumulating stale spreadsheet copies. This design provides the level of isolation needed for spreadsheet state while avoiding the resource cost of per-rollout servers or containers.

\paragraph{Scope of isolation.}
Filesystem workspace isolation is not intended to replace execution sandboxing for arbitrary code. Rather, it isolates the stateful spreadsheet artifacts that define each RL trajectory. When the agent invokes general Python code, execution is still delegated to the sandbox backend; when it needs faithful calculation, the workbook is sent through the asynchronous Excel API described in Appendix~\ref{sec:appendix_verifier_api}. The workspace design therefore composes with these external backends while keeping the core rollout interface simple enough to integrate with existing RL trainers.

\subsection{Cell-Level Evaluation Details}
\label{sec:appendix_cell_eval}
We compute outcome rewards by comparing the agent-edited workbook against an oracle workbook on the specified answer regions $M$. The comparison is performed on cached, post-recalculation cell values (i.e., formula cells are evaluated by their values). Before comparison, cell values are normalized following exactly how SpreadsheetBench did as follows:
\begin{itemize}
  \item Numbers are cast to float and rounded to 2 decimal places.
  \item Strings are parsed as numbers when possible and then rounded to 2 decimal places.
  \item \texttt{datetime} values are converted to Excel serial dates and rounded to the nearest day.
  \item \texttt{time} values are converted to \texttt{HH:MM}.
  \item Empty strings and \texttt{None} are treated as equivalent.
\end{itemize}
After normalization, a type mismatch counts as a mismatch; otherwise, values must match exactly.

\subsection{Verifier and Asynchronous Reward/Recalculation API}
\label{sec:appendix_verifier_api}
To score a model-edited workbook, the verifier must open the file in a faithful spreadsheet engine, trigger spreadsheet recalculation, save the resulting workbook (with Excel-cached formula values), and compare the answer-region cells against the oracle workbook. We deploy this verifier on a separate Windows CPU server because faithful execution requires Microsoft Excel.

\paragraph{Design constraints.}
Two operational constraints shaped the interface. First, end-to-end evaluation latency is highly variable: recalculation time depends on workbook size, the number of dependent formulas, volatile functions, and Excel runtime state. Second, in practice the Windows Excel server is commonly exposed behind a reverse proxy or tunnel, where long-lived HTTP requests are fragile. A purely synchronous ``compute-and-return'' endpoint would therefore (i) stall rollout workers for unpredictable durations and reduce GPU utilization during RL training, and (ii) be prone to dropped requests under network intermediaries when recalculation is slow.

\paragraph{Alternatives considered.}
\textbf{Synchronous HTTP reward computation} is the simplest design, but it requires holding a single request open while Excel performs recalculation and file I/O, making the system sensitive to timeouts and proxy disconnects, and turning tail latency into rollout-worker idle time.
\textbf{In-process Excel plugins/add-ins} can control Excel, but they couple each rollout trajectory to a persistent Excel session. This is expensive to scale, complicates resource accounting (Excel instances are memory-heavy), and makes failure recovery brittle under high concurrency.
\textbf{Non-Excel spreadsheet engines (e.g., LibreOffice)} avoid Windows and Excel, but we found them difficult to make reliable at the level of formula and function fidelity required for our benchmarks and real-world spreadsheets (e.g., gaps in function support and subtle behavioral mismatches). Since the goal of Spreadsheet-RL is to train agents against realistic Microsoft Excel semantics, we opt to keep Excel as the source of truth.
\textbf{Headless formula evaluation in Python} (e.g., \texttt{formulas}\footnote{\url{https://github.com/vinci1it2000/formulas}} and \texttt{xlcalculator}\footnote{\url{https://github.com/bradbase/xlcalculator}}) avoids Windows and Excel by interpreting formulas directly from \texttt{.xlsx}. However, this approach is ill-suited for Spreadsheet-RL: in our experiments it is substantially slower than native Excel recalculation on large dependency graphs (workbooks with many formula cells), and the supported Excel surface remains incomplete. For example, \texttt{formulas} reports 483/536 implemented functions (90.1\%), leaving gaps in categories such as database, cube, and web functions; \texttt{xlcalculator} explicitly does not support array/CSE formulas and documents known deviations for several functions (e.g., \texttt{LN}, \texttt{VLOOKUP}, and \texttt{YEARFRAC}). More broadly, even when a function is implemented, matching Excel's edge-case semantics (type coercion, numerical precision, rounding, and error propagation) is non-trivial; \texttt{xlcalculator} explicitly notes that keeping numeric behavior aligned with Excel may require a low-level numeric datatype beyond Python's built-in types. Such semantic mismatches can introduce subtle reward inconsistency relative to the target Excel execution environment.
Finally, slower evaluators also amplify end-to-end reward latency. Because RL optimization requires collecting outcome rewards before updating the policy, higher verifier latency backpressures rollout throughput and can reduce GPU utilization unless substantial buffering is introduced.

\paragraph{Overall architecture.}
We implement the verifier as an \emph{asynchronous job service} around real Excel execution. The API layer (FastAPI) accepts a workbook upload and immediately returns a job identifier; job state is persisted in a shared SQLite job store so multiple API workers can safely serve both submissions and polls. Background workers then claim queued jobs from the store and execute Excel-based recalculation in a bounded worker pool. This architecture separates \emph{short HTTP requests} from \emph{long-running Excel work}, while providing shared state for reliability and operational scaling.

\paragraph{Reward path.}
For reward computation, rollout workers call \texttt{POST /reward/submit} with a sample identifier (the thread directory) and the edited workbook. The service validates metadata (e.g., \texttt{answer\_position} and the corresponding oracle workbook), enqueues a reward job in SQLite, and returns a job identifier quickly. A background worker claims the job (transitioning \texttt{queued} $\rightarrow$ \texttt{running}), opens the workbook via Excel COM automation, triggers recalculation, and saves the processed workbook with updated cached values. The evaluator then compares the answer-region cells against the oracle workbook following \cref{sec:appendix_cell_eval}, and writes back both the scalar reward and a structured diagnostic message. The client obtains results by polling \texttt{GET /reward/result/\{job\_id\}} until the job reaches \texttt{done} or \texttt{error}. In our outcome-reward setting, the score is binary (match vs.\ mismatch), but the diagnostic message is essential for debugging, monitoring, and failure analysis at scale.

\paragraph{Recalculate path.}
Spreadsheet Gym also exposes a \texttt{recalculate\_and\_read} tool for intermediate self-verification during rollouts. This tool uses the same job machinery but returns the recalculated workbook rather than a reward number: clients submit a workbook to \texttt{POST /recalculate/submit}, the service enqueues a \texttt{kind=recalculate} job, and workers reuse the same bounded Excel pool to open, recalculate, and save the workbook. The client polls \texttt{GET /recalculate/result/\{job\_id\}} and receives the resulting \texttt{.xlsx} bytes. The rollout worker then reads the requested ranges from the recalculated workbook (using the Excel-cached values) and can optionally write back the updated workbook into the rollout workspace, ensuring that subsequent tool calls and the final episode output remain consistent with faithful Excel semantics.

\paragraph{Scaling and stability.}
To increase throughput, we maintain a pool of long-lived Excel instances to amortize cold-start overhead. Because Excel processes can accumulate memory or become unstable under sustained load, workers are recycled based on health signals (e.g., recycling when private bytes exceeds a threshold such as 4\,GB, or periodically after a fixed number of jobs). Finally, the worker pool is explicitly bounded to prevent overload and to keep the reward signal stable under high-concurrency RL training.

\paragraph{Capacity control and backpressure.}
The job service implements explicit backpressure to keep the system stable under bursty training loads. Concurrency is bounded by the size of the Excel worker pool, and submissions may be rejected when the pending queue exceeds a configured threshold, allowing rollout workers to retry with backoff rather than overwhelming the verifier. Polling is supported via lightweight status/result endpoints and can optionally use short server-side waits to reduce client polling frequency.

\paragraph{Failure handling and recovery.}
Each job is executed with a hard timeout and produces an explicit terminal state (\texttt{done} or \texttt{error}) recorded in SQLite, so transient failures or worker restarts do not corrupt job state. Long-lived Excel instances are proactively recycled to mitigate COM automation instability and memory growth, and stuck jobs can be detected and marked as failed to prevent indefinite queue buildup. These mechanisms are designed to keep reward delivery reliable at scale, rather than optimizing only for the median-case latency.

\paragraph{Semantic fidelity and determinism.}
Spreadsheet-RL treats Excel as the source of truth for evaluation: spreadsheets often rely on semantics that are difficult to reproduce faithfully outside Excel, including dynamic arrays and spill behavior, legacy array/CSE formulas, volatile functions, locale- and type-coercion corner cases, and numerically sensitive computations. Using real Excel recalculation ensures that both intermediate verification (\texttt{recalculate\_and\_read}) and terminal rewards are aligned with the semantics the agent is trained to operate under.

\paragraph{Data retention and privacy.}
The verifier operates on uploaded workbooks and oracle artifacts, but it does not require persisting these inputs long-term. Job state is stored as minimal metadata in SQLite, and uploaded/intermediate files can be aggressively cleaned up after completion (or retained briefly only for debugging), reducing both storage overhead and the risk of leaking sensitive spreadsheet contents.

\paragraph{Empirical throughput.}
In our 4B training runs, a single Windows CPU server with 32\,GB memory and a pool of four concurrent Excel instances sustained more than 20{,}000 reward/recalculation jobs in under 30 minutes (i.e., exceeding 11 reward/recalc jobs per second on average). This throughput is sufficient to keep the verifier from becoming the dominant bottleneck in practice and motivates the engineering emphasis on bounded concurrency, robustness, and short request/long job separation.

\subsection{GRPO Objective}
Concretely, for each input $[D_i, T]$, Spreadsheet Gym samples a group of response interactions (with the spreadsheet environments) of size $N$:
$\{ y_i \}_{i=1}^{N} \sim \pi_{\text{old}}(\cdot \mid D_i, T, G),$
from the old policy. The current policy $\pi_{\theta}$ is then updated by maximizing the objective:
\begin{equation}
\label{eq:grpo_condensed_fig}
\mathcal{L}_{\text{GRPO}}(\theta)
= \mathbb{E}\!\left[
\frac{1}{N}\sum_{i=1}^{N}
\min\!\left(
r_i(\theta)\widehat{A}_i,
\mathrm{clip}\!\left(r_i(\theta),1-\epsilon,1+\epsilon\right)\widehat{A}_i
\right)\right]
- \beta\,\mathrm{D}_{\mathrm{KL}}\!\left(\pi_{\theta}\|\pi_{\mathrm{old}}\right).
\end{equation}
\begin{equation}
\label{eq:grpo_ratio}
r_i(\theta)=
\frac{\pi_{\theta}(y_i \mid D_i,T;G)}
{\pi_{\mathrm{old}}(y_i \mid D_i,T;G)},
\end{equation}

\noindent where $\widehat{A}(y_i)$ is the group-relative advantage for response $y_i$, computed by normalizing outcome rewards within the group, and $\epsilon>0$ is the clip threshold. This formulation encourages exploration of diverse reasoning strategies, while maintaining stability and ensuring that the policy effectively improves relative to its peers within the sampled group. 

\subsection{Training Data Statistics}
\label{sec:appendix_train_data_stats}
\begin{figure}[H]
  \centering
  \begin{minipage}[c]{0.49\linewidth}
    \centering
    \begin{subfigure}[t]{\linewidth}
      \centering
      \includegraphics[width=\linewidth]{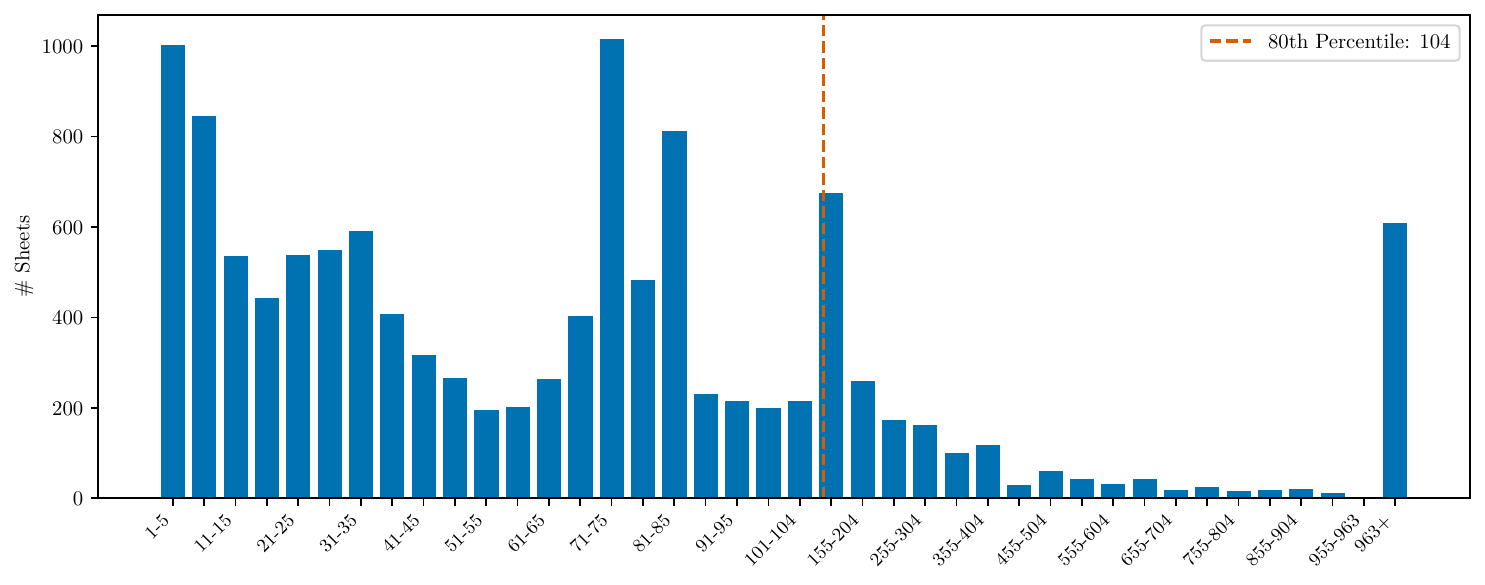}
      \caption{Row count distribution.}
      \label{fig:train-row-size-distribution}
    \end{subfigure}

    \vspace{0.6em}

    \begin{subfigure}[t]{\linewidth}
      \centering
      \includegraphics[width=\linewidth]{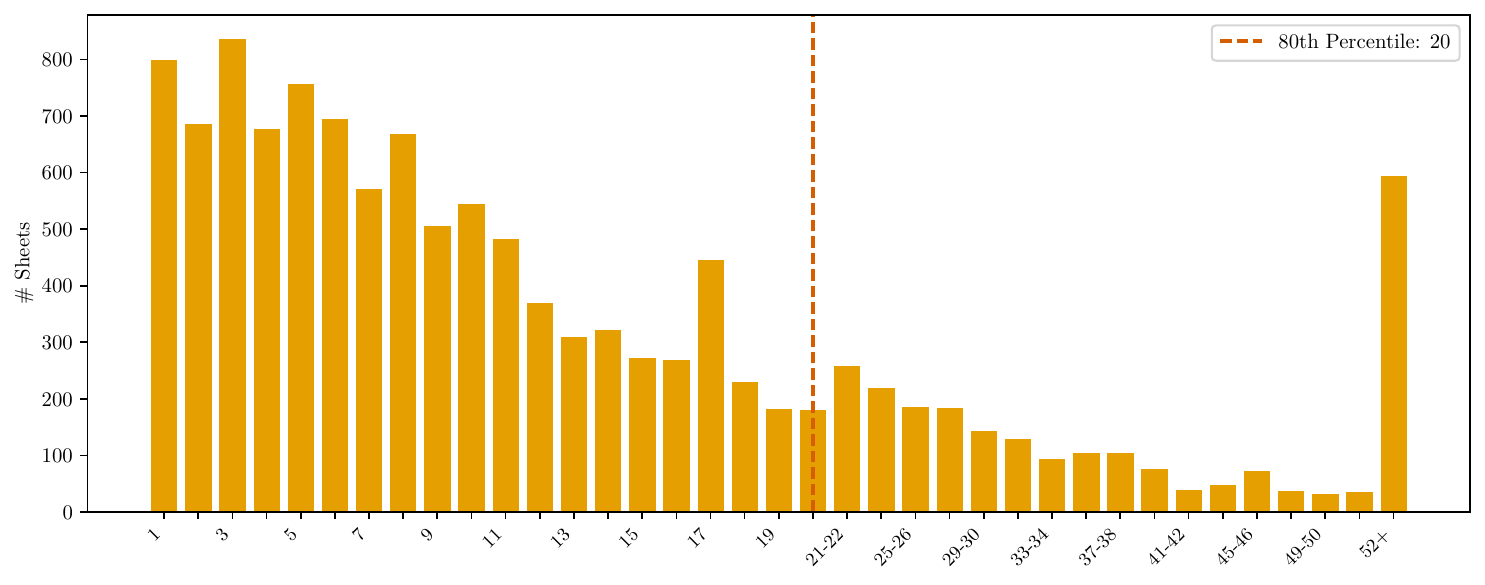}
      \caption{Column count distribution.}
      \label{fig:train-column-size-distribution}
    \end{subfigure}
  \end{minipage}
  \hfill
  \begin{minipage}[c]{0.49\linewidth}
    \centering
    \begin{subfigure}[t]{\linewidth}
      \centering
      \includegraphics[width=\linewidth]{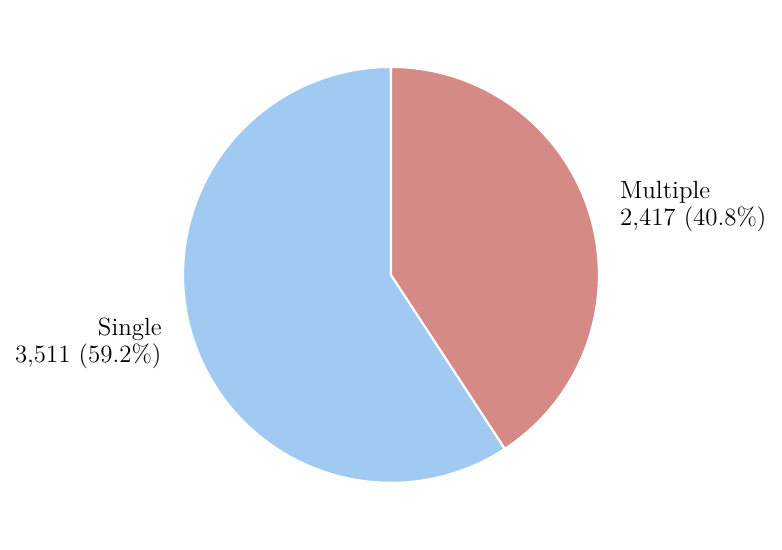}
      \caption{Worksheet count distribution.}
      \label{fig:train-sheet-proportion-distribution}
    \end{subfigure}
  \end{minipage}
  \caption{Training data spreadsheet size distributions.}
  \label{fig:train_data_stats}
\end{figure}

\begin{figure}[H]
  \centering
  \includegraphics[width=0.6\columnwidth]{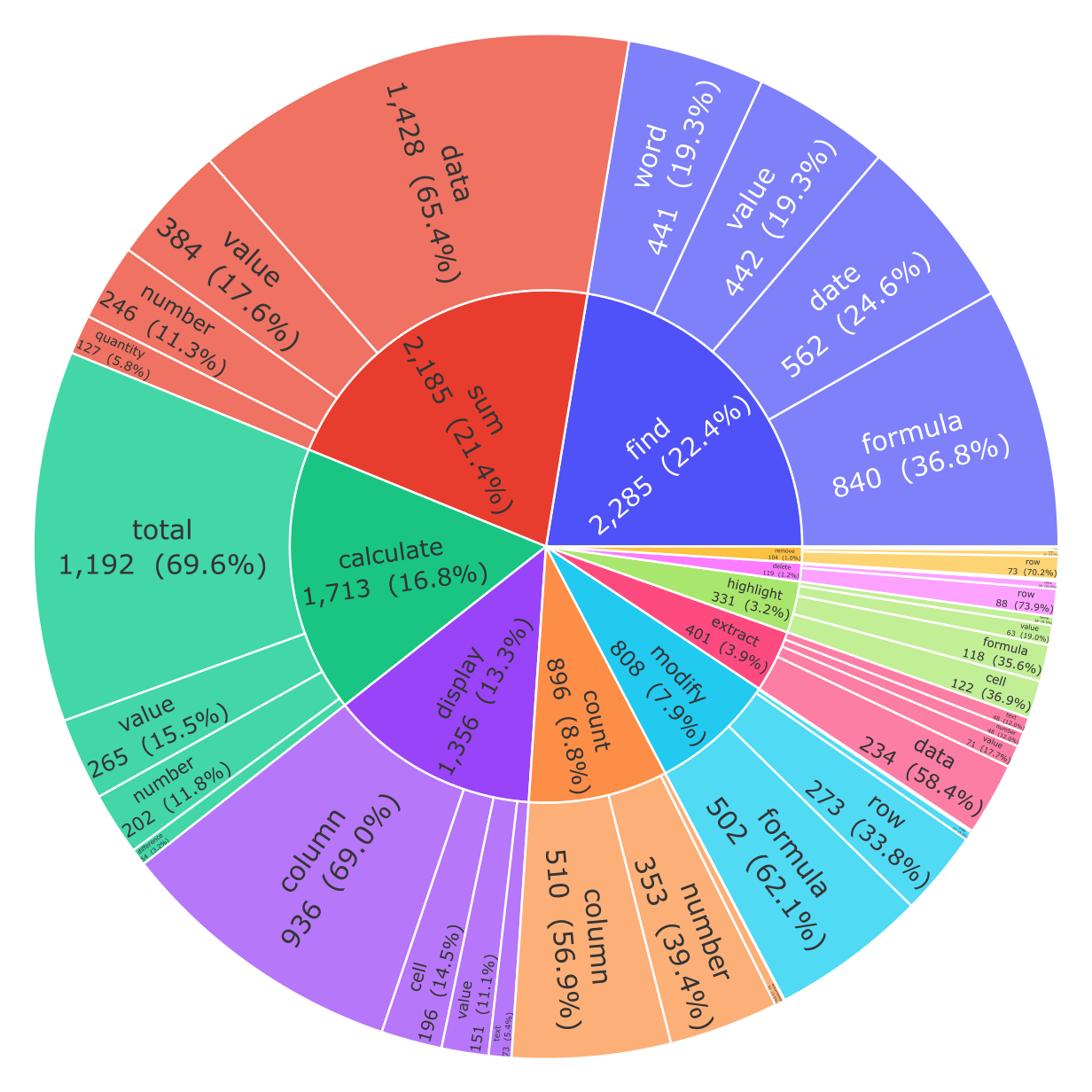}
  \caption{Sheet Operation distributions of RL Training Data.}
  \label{fig:verb-noun-phrase-distribution}
\end{figure}

\subsection{Training Config and Hyperparameters}
\label{sec:appendix_hparams}
For each rollout, we use a temperature of 0.6 with top-$p$ = 0.95 and top-$k$ = 20. Each training batch samples 16 rollouts per task over a set of 64 tasks, with a global batch size of 64. At each step, we perform two optimization updates using AdamW with a learning rate of $1\times10^{-6}$. We apply KL divergence with coefficient $0.001$ and enable dynamic batching for training efficiency. All training experiments are run on 1$\times$4 NVIDIA H100 GPUs; a full 4B training run takes approximately 40 hours of wall-clock time.
\begin{table}[H]
  \centering
  \small
  \caption{Training hyperparameters for the 4B run.}
  \label{tab:appendix_training_hparams}
  \setlength{\tabcolsep}{6pt}
  \begin{tabular}{@{}l p{0.72\linewidth}@{}}
    \toprule
    Hyperparameter & Value \\
    \midrule
    Base model & \texttt{Qwen/Qwen3-4B-Thinking-2507} \\
    Algorithm & GRPO; KL-regularized against a frozen reference model \\
    Training steps & 60 \\
    Prompt/response limits & 4{,}096 / 27{,}648 tokens \\
    Sampling (rollouts) & temperature 0.6; top-$p$ 0.95; top-$k$ 20 \\
    Batching & 64 prompts/step; 16 rollouts/prompt ($N=16$); 1{,}024 rollouts/step \\
    Multi-turn caps & max assistant turns 20; max user turns 20; max tool-response length 8{,}192 \\
    Interaction protocol & inspect-modify-verify spreadsheet harness prompt;\\
    Optimizer & AdamW; learning rate $1\times 10^{-6}$; weight decay 0.01; $(\beta_1,\beta_2)=(0.9,0.999)$; grad clip 1.0 \\
    KL loss & low-var KL; coefficient 0.001 \\
    Actor update batching & mini-batch 32; dynamic batch sizing enabled \\
    \bottomrule
  \end{tabular}
\end{table}


\end{document}